\documentclass[fleqn,10pt]{wlscirep}
\usepackage[utf8]{inputenc}
\usepackage[T1]{fontenc}
\usepackage{tikz}
\usepackage{subcaption}
\usepackage{multirow}
\usetikzlibrary{shapes.geometric, arrows.meta, calc, positioning, shadows}
\title{Weakly supervised multimodal segmentation of acoustic borehole images with depth-aware cross-attention}
\author[1,2]{Jose Luis Lima de Jesus Silva}
\affil[1]{Federal University of Bahia, Institute of Geosciences, Department of Geophysics, Salvador, BA 40170-115, Brazil}
\affil[2]{Grupo de Estudos e Aplicação de Inteligência Artificial em Geofísica (GAIA), Federal University of Bahia, Salvador, BA 40170-115, Brazil}

\keywords{weak supervision, multimodal segmentation, depth-aware cross-attention, confidence-aware fusion, acoustic borehole images, petrophysical logs}

\begin{abstract}
Acoustic borehole images provide high-resolution borehole-wall structure, but large-scale interpretation remains difficult because dense expert annotations are rarely available and subsurface information is intrinsically multimodal. The challenge is developing weakly supervised methods combining two-dimensional image texture with depth-aligned one-dimensional well-logs. Here, we introduce a weakly supervised multimodal segmentation framework that refines threshold-guided pseudo-labels through learned models. This preserves the annotation-free character of classical thresholding and clustering workflows while extending them with denoising, confidence-aware pseudo-supervision, and physically structured fusion. We establish that threshold-guided learned refinement provides the most robust improvement over raw thresholding, denoised thresholding, and latent clustering baselines. Multimodal performance depends strongly on fusion strategy: direct concatenation provides limited gains, whereas depth-aware cross-attention, gated fusion, and confidence-aware modulation substantially improve agreement with the weak supervisory reference. The strongest model, confidence-gated depth-aware cross-attention (CG-DCA), consistently outperforms threshold-based, image-only, and earlier multimodal baselines. Targeted ablations show its advantage depends specifically on confidence-aware fusion and structured local depth interaction rather than model complexity alone. Cross-well analyses confirm this performance is broadly stable. These results establish a practical, scalable framework for annotation-free segmentation, showing multimodal improvement is maximized when auxiliary logs are incorporated selectively and depth-aware.
\end{abstract}
\begin{document}

\flushbottom
\maketitle
%
%
\thispagestyle{empty}

\section*{Introduction}

High-resolution borehole acoustic image logs provide a detailed, two-dimensional view of structures within the rock around a well, revealing features such as fractures, breakouts, and enlargements caused by stress, bedding-related textures and patterns from layered sediment, and other geological attributes that are hard to capture with one-dimensional well logs alone \cite{prensky1999advances, luthi2001geological}. In petroleum engineering and geoscience workflows, these images are interpreted to support reservoir (rock containing hydrocarbons) and geomechanical (rock stress and strength) analysis, completion design, and subsurface risk assessment \cite{dias2020automatic}. However, manual interpretation is labor-intensive, subjective, and difficult to scale across large well inventories \cite{peveraro2021role}. As a result, we need automated methods that are accurate, robust, interpretable, and capable of leveraging complementary multimodal well data (data from different well measurement tools) \cite{hall2016facies}.

A major challenge is that acoustic images and conventional logs have fundamentally different structures. Acoustic images are two-dimensional datasets that show both the depth and the direction (bearing) around the borehole and display detailed patterns on the borehole wall (spatial texture). In contrast, conventional well logs, such as gamma ray (which measures natural radioactivity), density (the compactness of the formation), neutron porosity (the amount of hydrogen, indicating porosity), sonic travel time (the speed of sound through the formation), and resistivity (resistance to electrical current flow), are one-dimensional measurements recorded only by depth \cite{lai2024application}. Both data types are aligned along the depth axis, but they capture different physical properties of the subsurface formation \cite{kong2023segmentation}. The acoustic image shows patterns and variations on the borehole wall at each depth, while customary logs provide information about the material properties sampled vertically. Successful data fusion has to account for this difference in geometry, rather than treating all channels as if they provide the same type of input \cite{he2021multimodal}.

Another major obstacle is the scarcity of reliable dense annotations. Pixel-level or sample-level labeling of borehole images is expensive and requires expert knowledge \cite{ye1998specialized}. In practice, many interpretation workflows rely on threshold-based heuristics, denoising steps, and manual refinement rather than on large, curated, tagged datasets \cite{anovitz2015characterization}. This motivates weakly supervised and self-refining approaches, which can exploit structure already present in the data without requiring fully supervised training \cite{zhou2018brief}. The challenge is to turn noisy pseudo-labels into useful supervision while preventing the model from overfitting artifacts or spurious multimodal correlations. Classical annotation-free approaches such as thresholding, clustering, and self-organizing maps can reveal coarse structure in borehole images \cite{hasan2023lithofacies, keretchashvili2025unsupervised}, but they generally do not have explicit mechanisms for confidence-aware refinement, depth-aligned multimodal interaction, and selective control over auxiliary information. As a result, they are useful for exploratory partitioning but frequently remain limited in spatial coherence, multimodal interpretability, and dependability across intervals.

The lack of dense manual labels is not a temporary inconvenience but a defining condition of the problem. Our framework was designed for this realistic regime of annotation-free analysis, preserving the practical advantage of no dense expert labels. At the same time, it introduces stronger refinement mechanisms to improve structural unity and multimodal interpretability.
Recent multimodal learning research in petrophysics has shown that relationships between borehole image logs and conventional logs can be learned through self-organizing maps, heteroassociative mappings, and self-supervised regenerative frameworks. These enable cross-modal representation learning, probabilistic multi-output prediction from heterogeneous 1D and 2D inputs \cite{lin2023deep}, and the generation or regeneration of data \cite{Frota2024IJCNN,Frota2024CBA,Frota2026Geoenergy,Frota2026Neurocomputing}. However, these approaches are mainly for generation, prediction \cite{lin2024deep}, or representation learning \cite{qi2025wlfm}, not dense interpretation problems. This work builds on previous research by treating multimodal petrophysical learning as a weakly supervised segmentation task applied to acoustic borehole images, aligning increasingly tightly with recent semi-supervised strategies that reduce annotation needs in similar segmentation tasks \cite{da2025semi}. We develop a geometry-aware fusion strategy where images are treated as depth-azimuth fields, whereas conventional logs serve as depth-indexed contextual signals. Their interaction is mediated through local depth-aware cross-attention \cite{chen2025geological}, learned gating, and confidence-aware fusion. This approach shifts the focus from cross-modal generation to annotation-efficient, spatially explicit interpretation \cite{shen2025fracloggpt}, facilitating selective, reliability-aware multimodal refinement of borehole-image structure.

Accordingly, our contribution is to move from multimodal petrophysical generation and representation learning toward weakly supervised, dense borehole-image interpretation with geometry-aware and confidence-aware fusion.

In this work, we deal with these challenges using a multimodal, weakly supervised framework to segment and refine acoustic borehole intervals. Our pipeline starts with threshold-derived pseudo-supervision, in which approximate training labels are constructed from raw and denoised acoustic images through global and local flexible thresholding. We progressively evaluate stronger refiners with increasing multimodal structure \cite{valentin2019deep}. Our framework first considers image-only baselines and direct multimodal concatenation from multiple data sources. Then we introduce depth-aware cross-attention, a method that explicitly links each row in the image, representing a specific measured depth in the borehole, to conventional log measurements such as electrical resistivity or sonic travel time. The aim is to determine which log-derived petrophysical context is most relevant to the interpretation of the visual structure observed at a given image depth.

To improve robustness, we also investigate adaptive fusion mechanisms. These mechanisms flexibly combine information from different sources and modulate the contribution of conventional logs. Conventional logs are primarily recorded data from sensors or systems, rather than data assumed to be of consistent benefit. In particular, we examine gated variants that use learned thresholds to control information flow. We also examine confidence-aware variants that adjust fusion based on the system's certainty. The last attributes tune multimodal interaction based on the estimated reliability of the pseudo-supervision. Pseudo-supervision refers to labels or annotations from non-expert or automated sources, as well as local segmentation context, or specific data regions \cite{jseluis2023wavelet, wu2025beyond}. It is important to note that multimodal information is not always useful across all intervals because it can clarify an ambiguous structure. Therefore, a successful model must learn not only how to fuse the modalities, but also when and where to trust the additional information.

To ground our evaluation in a realistic, operationally relevant setting, this study leverages the Wellbore Acoustic Image Database (WAID), an open-source repository released by PETROBRAS. The WAID dataset provides high-resolution acoustic amplitude images alongside a comprehensive suite of depth-aligned conventional open-hole logs—including caliper, gamma ray, bulk density, neutron porosity, sonic slowness, and resistivity—from five distinct wells located in a highly heterogeneous Brazilian carbonate pre-salt reservoir. To protect confidentiality, these wells have been pseudonymized after endangered species (Antilope-25, Antilope-37, Botorosa-47, Coala-88, and Tatu-22). Although the dataset was released explicitly to foster artificial intelligence innovation in petrophysics, we intentionally operate in the unannotated regime to maintain our core focus on developing new approaches to advance weakly supervised, annotation-free interpretation techniques. This choice makes WAID an ideal testbed for evaluating our framework under the practical constraints faced in scalable, real-world workflows. While recent literature has successfully utilized WAID to pioneer self-supervised regenerative learning, cross-modal representations, and bidirectional probabilistic log generation \cite{Frota2024IJCNN, Frota2024CBA, Frota2026Geoenergy, Frota2026Neurocomputing}, our work represents a novel expansion of its utility. We shift the application of this dataset away from purely generative tasks and toward dense, geometry-aware, and confidence-modulated structural interpretation.

We analyze benchmarks to investigate the benefits of multimodal attention. This is assessed by comparing performance among wells, interval types from distinct geological or data-defined intervals, and representative interval conditions, with varying levels of segmentation difficulty. We relate improvements to factors such as the model’s prediction uncertainty, reconstruction error, and segmentation morphology, including transition density and structural fragmentation. This approach elevates our contribution from a purely empirical benchmark to a more scientific analysis of multimodal behavior in borehole interpretation. Such analysis is especially relevant in production contexts, where decision-makers need to know when a method is reliable and when manual review is needed. Our framework additionally strengthens the weak supervisory signal. We compare threshold-based pseudo-labeling, image-only refinement, direct multimodal concatenation, and the proposed attention-based multimodal refiners. Performance is summarized mainly through permutation-invariant agreement with the pseudo-label reference, which measures agreement with the weak supervisory signal while taking into account class-label ambiguity. The cross-well benchmarking and targeted ablation are then used to identify both the strongest model and the mechanisms responsible for its gains. We observe that agreement increases from 0.6002 for raw thresholding to 0.7456 after denoising-based threshold guidance. Across the full cross-well benchmark, the strongest model, confidence-gated depth-aware cross-attention (CG-DCA), which combines local depth-aware cross-attention with learned gating and confidence-aware fusion, reaches a mean agreement of 0.8571. It outperforms direct multimodal concatenation, which fuses image and log information through early channel-wise stacking (0.7518), the image-only refiner, which relies only on acoustic image information (0.7339), and ungated depth-aware cross-attention (DCA), which uses depth-local cross-attention without selective confidence-aware fusion (0.8044). Targeted ablation further shows that this gain is mechanistically meaningful rather than incidental. Full CG-DCA achieves a mean agreement of 0.9172 on the ablation subset, dropping to 0.8904 when confidence-aware fusion is removed.

\section*{Results}
\subsection{Threshold-guided refinement and multimodal fusion}
\label{subsec:initial_results_antilope25}

\begin{figure*}[htbp]
\centering
\includegraphics[width=\textwidth]{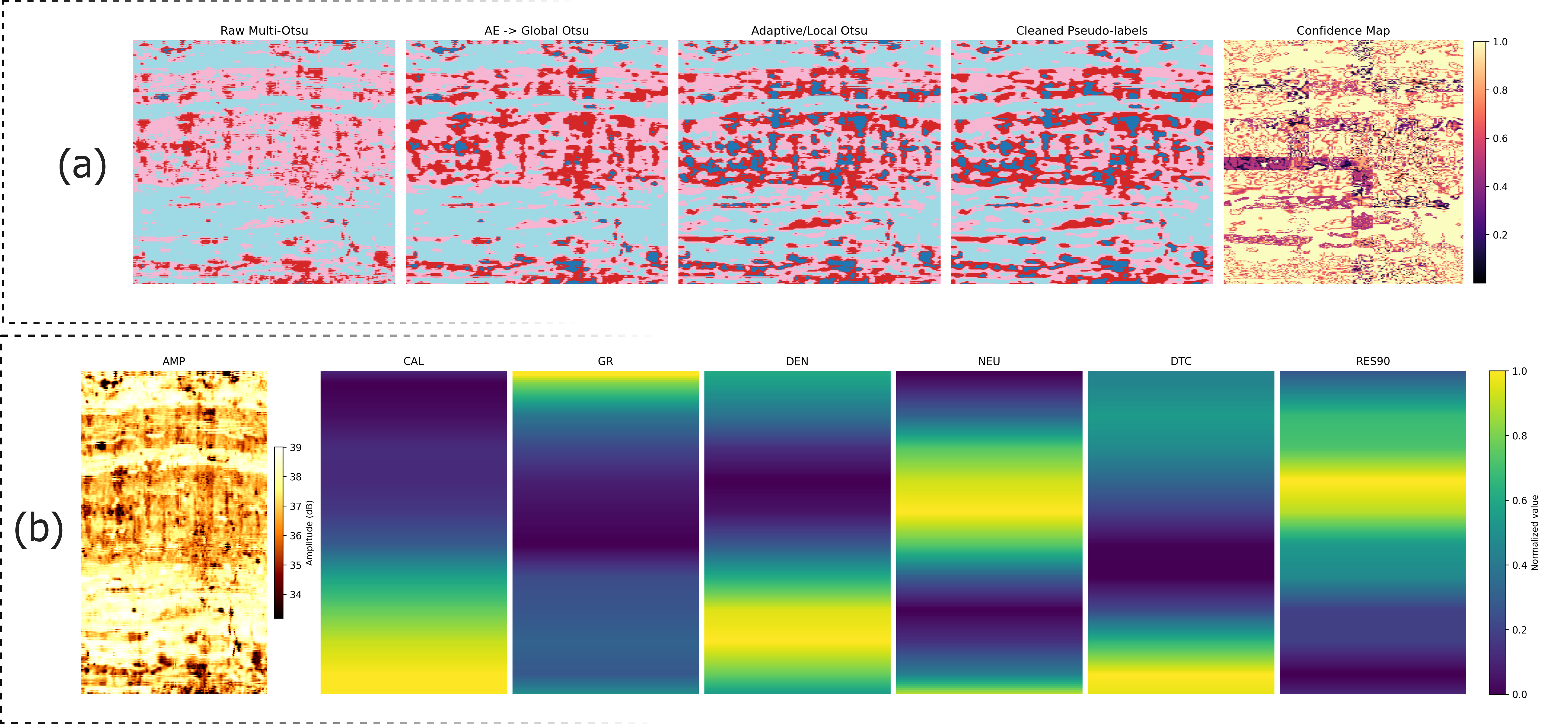}
\caption{Construction of the weak supervisory signal and aligned multimodal inputs for the initial Antilope25 interval. \textbf{(a)} From left to right: raw global Multi-Otsu thresholding, denoised global Multi-Otsu thresholding, adaptive/local thresholding, cleaned pseudo-labels, and the corresponding confidence map. \textbf{(b)} From left to right: acoustic image amplitude (AMP), caliper (CAL), gamma ray (GR), bulk density (DEN), neutron porosity (NEU), compressional slowness (DTC), and deep resistivity (RES90). The log channels were interpolated to the image depth grid, normalized, and laterally replicated into image-shaped auxiliary channels.}
\label{fig:threshold_guidance}
\end{figure*}
We first assessed whether a representative interval from the Antilope25 well could be partitioned into a coherent structural map using the borehole acoustic image, and whether this partition could be improved through incorporating depth-aligned conventional well logs. Because manual pixel-level labels were not available, the analysis was conducted under limited supervision \cite{qi2025wlfm}. In this setting, the learned models were trained against automatically generated pseudo-labels rather than expert geological annotations. Three objects should therefore be distinguished throughout this first analysis: the input acoustic image, the pseudo-label map used as a provisional training target, and the refined segmentation predicted by the neural network. A four-class partition was adopted here as a reproducible segmentation scale for the method, rather than as evidence that the interval contains exactly four geological facies \cite{kong2023segmentation}. A permutation-invariant metric (see Methods) serves as the main benchmark, as arbitrary class numbering is inherent to these partitions. This metric scores the optimal match against the pseudo-label map and measures internal consistency rather than externally validated geological ground truth. We also compute structural diagnostics to explicitly map geometric changes in the class boundaries.

Figure~\ref{fig:threshold_guidance}a summarizes how the provisional targets were obtained. When global Multi-Otsu thresholding\cite{otsu1979threshold} is applied directly to the raw image, the main contrast pattern is recovered, but the result is still affected by local noise. Using the autoencoder-denoised image makes this large-scale partition more stable. By applying local adaptive thresholding to resolve depth-varying contrast, we obtain a regularized final pseudo-label map and a paired confidence map that systematically assigns lower certainty to ambiguous boundaries. To construct the multimodal input (Figure~\ref{fig:threshold_guidance}b), we have used conventional well logs including caliper (CAL), gamma ray (GR), bulk density (DEN), neutron porosity (NEU), compressional slowness (DTC), and deep resistivity (RES90). These logs were interpolated to the image depth grid, independently normalized, and laterally replicated. These channels provide a structured and depth-dependent petrophysical context that complements the two-dimensional acoustic texture (see Supplementary Table~S1 for descriptive statistics). Note that the lateral continuity of the aligned log channels in Figure~\ref{fig:threshold_guidance}b is merely an artifact of azimuthal replication, since the true physical variation occurs exclusively along the vertical depth axis.

\begin{figure*}[htbp]
\centering
\includegraphics[width=\textwidth]{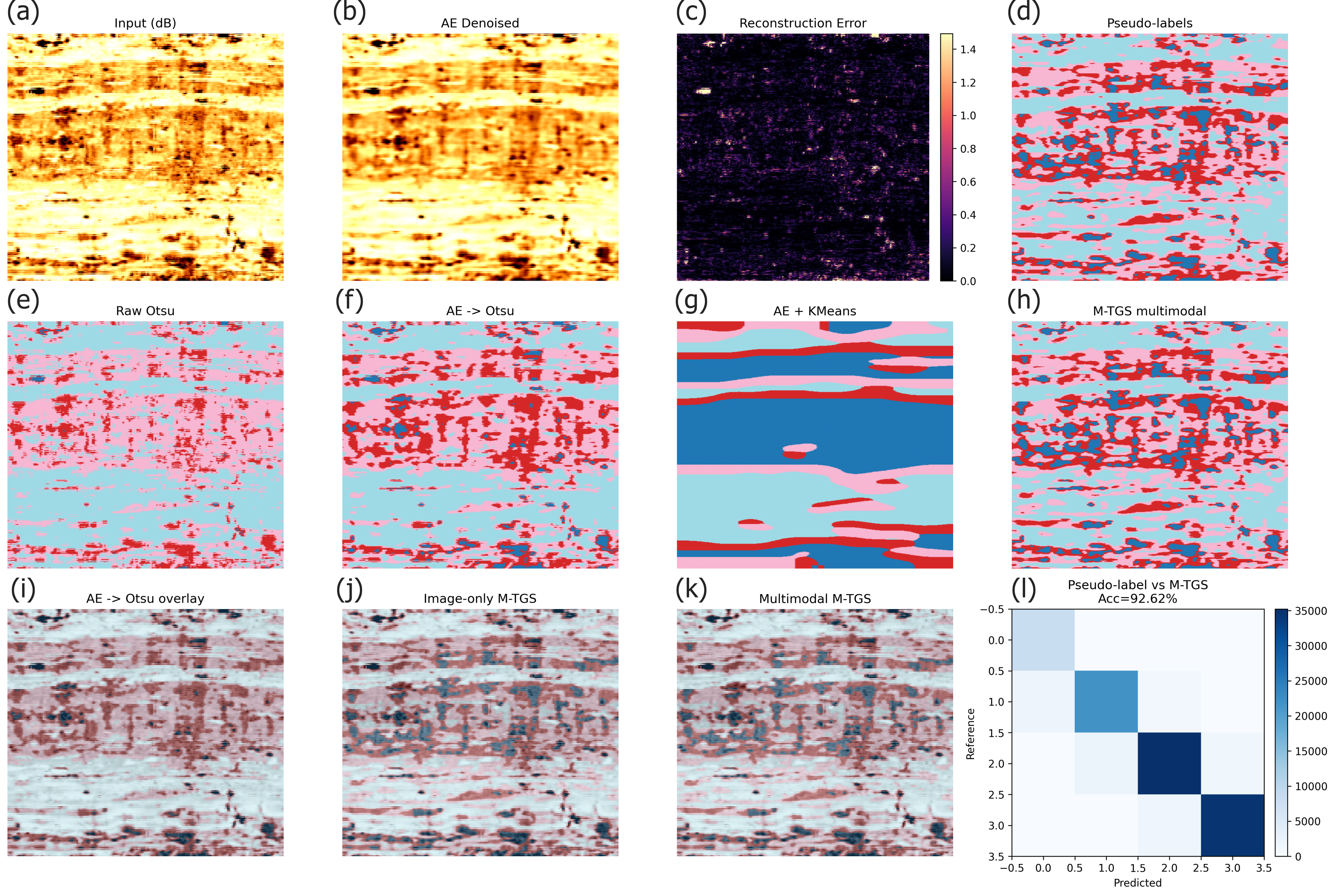}
\caption{Initial multimodal segmentation findings for the representative Antilope25 interval. \textbf{(a)} Input acoustic borehole image in decibel units. \textbf{(b)} Autoencoder-denoised reconstruction. \textbf{(c)} Reconstruction error map. \textbf{(d)} Cleaned pseudo-label map. \textbf{(e)} Raw global Multi-Otsu thresholding. \textbf{(f)} Global Multi-Otsu thresholding applied to the denoised image. \textbf{(g)} AE + KMeans baseline. \textbf{(h)} Final multimodal M-TGS segmentation using the denoised acoustic image and aligned auxiliary logs. \textbf{(i)} Overlay of the AE $\rightarrow$ Otsu segmentation. \textbf{(j)} Overlay of the image-only M-TGS segmentation. \textbf{(k)} Overlay of the multimodal M-TGS segmentation. \textbf{(l)} Confusion matrix comparing the pseudo-label reference with the multimodal M-TGS segmentation.}
\label{fig:publication_panel}
\end{figure*}

The full progression from raw input to refined segmentation is captured in Figure~\ref{fig:publication_panel}. The original acoustic image Figure~\ref{fig:publication_panel}a is characterized by laterally continuous yet locally irregular banding. By passing this through the autoencoder Figure~\ref{fig:publication_panel}b, fine-scale textural noise is attenuated while the primary structural organization remains intact. The reconstruction error map Figure~\ref{fig:publication_panel}c shows that this smoothing is highly selective and due to irregular anomalies and sharp lower horizons rather than washing out major boundaries. This controlled denoising yields a pseudo-label map Figure~\ref{fig:publication_panel}d that traces the dominant lateral architecture with far greater spatial coherence than naive thresholding.

The raw global thresholding in Figure~\ref{fig:publication_panel}e remains highly fragmented, reaching a permutation-invariant agreement (see Methods section) of 0.6002 with the pseudo-label reference. This metric measures the fraction of pixels that agree after optimal matching of class identities between the two segmentations. However, the denoised thresholding in Figure~\ref{fig:publication_panel}f) improves spatial continuity and raises agreement to 0.7456. Because the reference is itself a pseudo-label map rather than an expert geological ground truth. Therefore, these values should be interpreted as measures of internal consistency rather than absolute geological accuracy. Furthermore, the latent-clustering reference from  the method AE + KMeans (see Methods section), Figure~\ref{fig:publication_panel}g produces blocky, artificial partitions and reaches only 0.4212 agreement, indicating that learned features alone cannot recover coherent structure without spatial refinement (quantitative results in Supplementary Table~S2). This shows that the gain observed later in the pipeline is not a generic consequence of using learned features, but depends on threshold-guided spatial refinement.

The proposed multimodal Threshold-Guided Segmentation (M-TGS) model (see in Methods) overcomes these drawbacks by learning to improve the provisional target, weighted by the confidence map to ignore unstable boundaries. The final multimodal M-TGS segmentation (Figure~\ref{fig:publication_panel}h) successfully consolidates the horizontal bands and internal anomalies, achieving the highest overall agreement of 0.9262 (compared to 0.8938 for the image-only refiner). Here, agreement denotes permutation-invariant agreement with the pseudo-label reference, which means that the fraction of pixels assigned to the same class after optimal matching of class identities between the compared segmentations (see Methods). Relative to denoised thresholding, the image-only refiner improves agreement by 0.1482 and the multimodal model by 0.1806, while the multimodal gain over the image-only baseline is 0.0324. The overlay visualizations confirm this structural precision as compared to the denoised Otsu reference (Figure~\ref{fig:publication_panel}i) and the image-only refiner (Figure~\ref{fig:publication_panel}j), the multimodal overlay (Figure~\ref{fig:publication_panel}k) most accurately corresponds to the visible boundaries of the authentic image, particularly in the ambiguous central regions. The confusion matrix, which counts how many pixels assigned to each pseudo-label class are mapped to each predicted class after optimal class matching (Figure~\ref{fig:publication_panel}), shows a strong diagonal pattern. Therefore, high agreement is distributed across the full four-class partition rather than collapsed into a single dominant background class.

We broadened our tests to include cross-well screening to determine whether these improvements worked across different rock layers. We found that the refined method consistently outperformed older methods. The image-only refiner matched the pseudo-label reference with a mean score of 0.8349, beating raw Multi-Otsu thresholding (0.6670), denoised thresholding (0.7223), and unsupervised latent clustering AE + KMeans with a score of 0.5201 (see Supplementary Table S3). However, our screening also revealed that the results from concatenated depth-aligned conventional logs with the acoustic image (multimodal integration) varied widely across geological intervals (slices). This approach outperformed the image-only refiner, but overall it surpassed the refiner in only 7 of 15 slices, resulting in a negligible median difference. Such variation suggests that unselective injection of petrophysical logs can dilute strong image-derived features rather than universally improve segmentation.

We isolated three morphologically diverse intervals for extended optimization to determine exactly when and why multimodal fusion succeeds or fails (Supplementary Table~S4). Selected across different wells and depth regimes, these cases embody distinct structural challenges. A laterally banded interval from Botorosa47 provides an ideal environment for multimodal integration. Conversely, a deep, vertically columnar section of Antilope25 is resolved so effectively by the image-only model that simple multimodal fusion offers no advantage. Finally, a mid-depth Antilope25 interval featuring a compact, localized anomaly serves as a testing case for naive multimodal concatenation.

To evaluate these varied geometries, we apply a consistent visual analysis to each interval. We compare the raw acoustic input (panel a) against its structure-preserving denoised reconstruction (panel b). Discrepancies between the two are captured by the log-compressed reconstruction error (panel c), while a pixel-wise confidence map (panel d) identifies ambiguous transitions to be down-weighted during optimization. Next, we benchmark the raw thresholding (panel e), denoised thresholding (panel f), and AE + KMeans clustering, directly against the learned image-only (panel g) and multimodal (panel h) refiners. This framework was used to investigate how the laterally banded, vertically columnar, and highly localized structures can shed light on the physical conditions under which auxiliary logs aid or disrupt spatial segmentation.

\subsection{Botorosa47 multimodal case}
\label{subsec:botorosa47_detailed}
\begin{figure}[htbp]
\centering
\includegraphics[width=.9\textwidth]{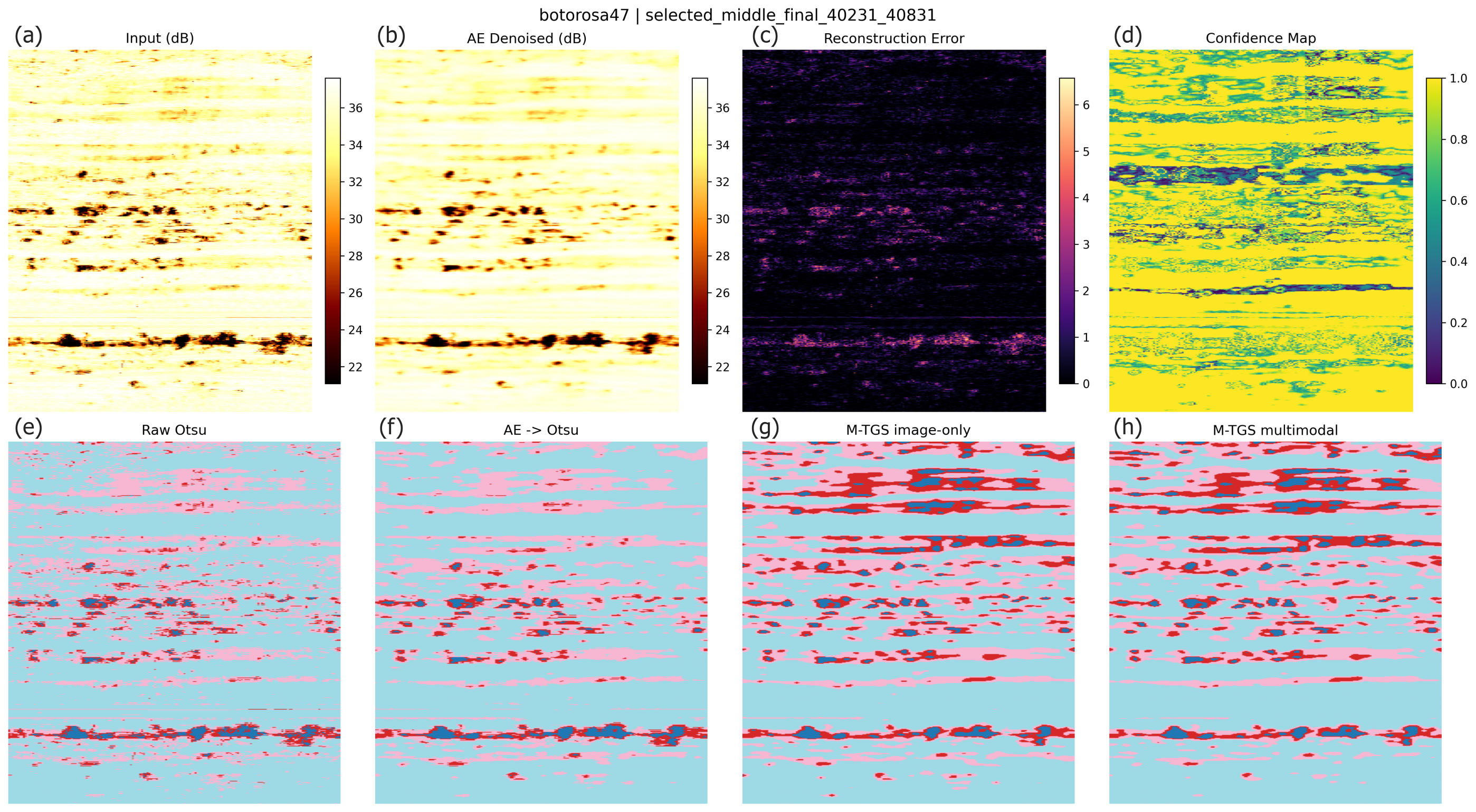}
\caption{Botorosa47 multimodal case. Top row: \textbf{(a)} input acoustic image, \textbf{(b)} autoencoder-denoised reconstruction, \textbf{(c)} reconstruction error, and \textbf{(d)} confidence map. Bottom row: \textbf{(e)} raw thresholding, \textbf{(f)} denoised thresholding, \textbf{(g)} image-only refined segmentation, and \textbf{(h)} multimodal refined segmentation. The interval is characterized by laterally continuous banded structure. Both refined models produce segmentation maps that are markedly more coherent than the thresholding baselines, while the multimodal refiner yields the strongest overall result, achieving an agreement of 0.9133 with the pseudo-label reference.}
\label{fig:botorosa47_heavy}
\end{figure}

Figure~\ref{fig:botorosa47_heavy} illustrates the clearest multimodal-positive case within a morphologically legible interval from the Botorosa47 well. The input acoustic image, Figure~\ref{fig:botorosa47_heavy}a, exhibits laterally continuous horizontal banding interrupted by localized, low-intensity (darker) anomalies. Furthermore, the autoencoder preserves this dominant layering while attenuating high-frequency noise in Figure~\ref{fig:botorosa47_heavy}b. The reconstruction error (log-compressed) shown in Figure~\ref{fig:botorosa47_heavy}c is concentrated around complex structural boundaries and localized anomalies. The latter suggests that the denoiser selectively smooths local complexity. In Figure~\ref{fig:botorosa47_heavy}d, we observe high reliability to broad and coherent bands, and down-weighting ambiguous transition zones, therefore, providing a structurally aware weakly supervised training signal. Before attempting to classify the rock formations or use the 1D well logs, we first have to clean the raw image and identify which parts we can actually trust.

Classical global thresholding in Figure~\ref{fig:botorosa47_heavy}e is overly sensitive to local variability, producing a highly fragmented partition that improves only modestly from an agreement of 0.7309 to 0.7595 after denoising Figure~\ref{fig:botorosa47_heavy}f. To overcome this limitation, we apply the M-TGS refiner, which is optimized using the pseudo-label map as a provisional target and spatially regulated by the confidence map to ignore unstable boundaries. The image-only M-TGS refiner Figure~\ref{fig:botorosa47_heavy}g drastically reorganizes the fragmented baseline into continuous bands, increasing agreement to 0.8664. The multimodal M-TGS refiner Figure~\ref{fig:botorosa47_heavy}h provides a good morphology and overall agreement of 0.9133. By incorporating petrophysical logs with the acoustic image, the multimodal network successfully stabilizes class assignments across laterally continuous but locally ambiguous bands. Therefore, auxiliary 1D context is highly beneficial for resolving continuous horizontal architectures.

The quantitative progression in this interval is consistent with the visual interpretation, which shows that agreement increases from 0.7309 for raw thresholding and 0.7595 after denoising to 0.8664 for the image-only refiner and 0.9133 for the multimodal refiner. Therefore, the aligned auxiliary logs can contribute an additional improvement when the interval is organized as a laterally continuous but locally ambiguous banded structure.

\subsection{Strong refinement in a vertically elongated morphology}
\label{subsec:antilope25_bottom_detailed}
\begin{figure}[htbp]
\centering
\includegraphics[width=.9\textwidth]{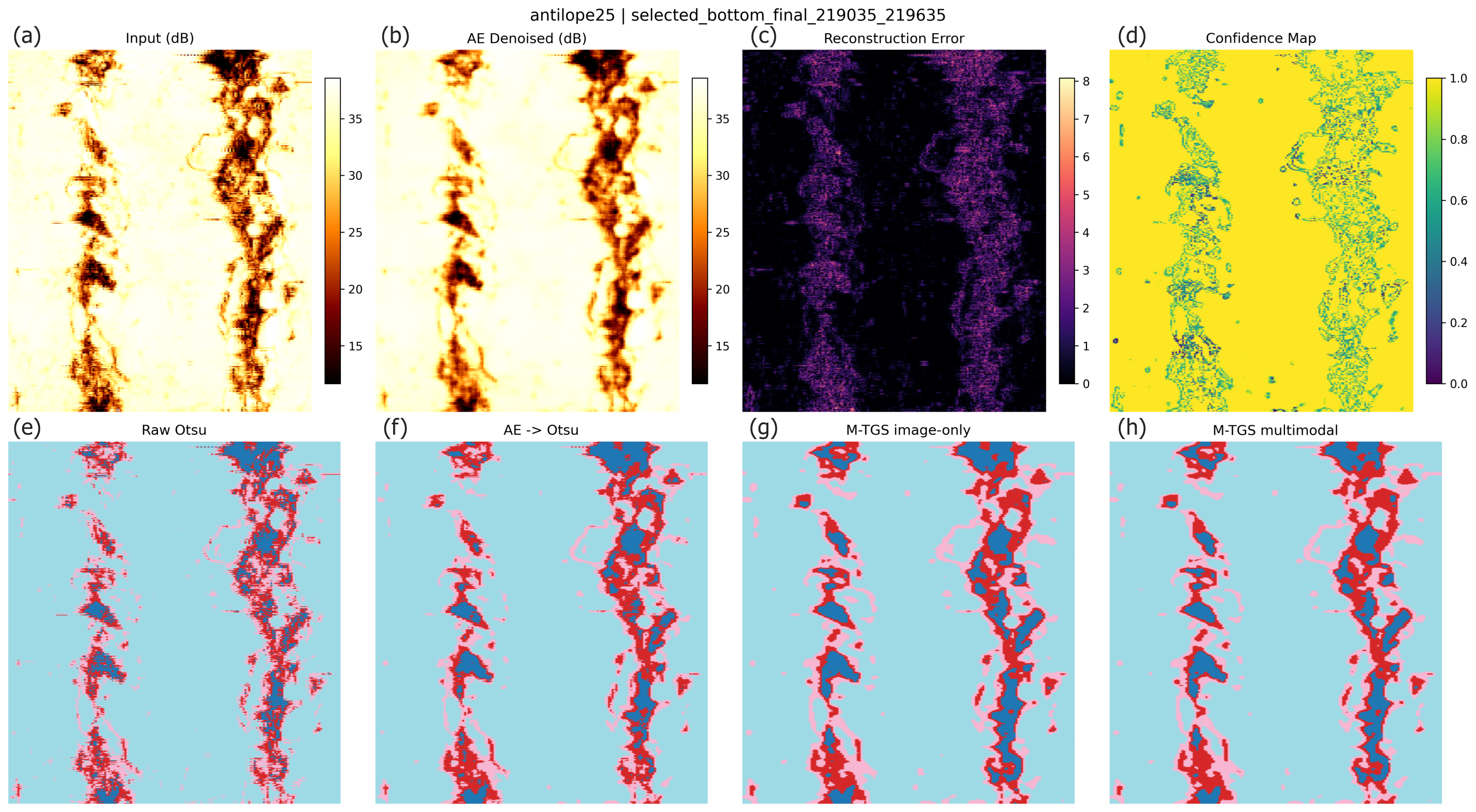}
\caption{Heavy-rerun interval from Antilope25 displaying robust refinement in a distinct morphology. Top row: \textbf{(a)} input acoustic image, \textbf{(b)} autoencoder-denoised reconstruction, \textbf{(c)} reconstruction error, and \textbf{(d)} confidence map. Bottom row: \textbf{(e)} raw thresholding, \textbf{(f)} denoised thresholding, \textbf{(g)} image-only refined segmentation, and \textbf{(h)} multimodal refined segmentation. The interval is dominated by vertically elongated structures. Both optimized models substantially outperform the thresholding references.}
\label{fig:antilope25_bottom_heavy}
\end{figure}
Figure~\ref{fig:antilope25_bottom_heavy} evaluates the refinement framework in a second representative interval from the Antilope25 well, which is morphologically dominated by vertically elongated, columnar structures rather than lateral bands. The input image Figure~\ref{fig:antilope25_bottom_heavy}a presents a highly distinctive structural signal with strong contrast against the background. The autoencoder successfully preserves this vertical geometry, as shown in Figure~\ref{fig:antilope25_bottom_heavy}b, concentrating reconstruction error primarily along the irregular edges and local protrusions of the columns (Figure~\ref{fig:antilope25_bottom_heavy}c). Figure~\ref{fig:antilope25_bottom_heavy}d highlights higher-confidence interior regions and lower-confidence structural margins. The latter suggests the successful preservation of supervision over the broader geometric bodies.

While direct thresholding captures the broad location of these structures, the raw partition remains noisy and over-segmented in Figure~\ref{fig:antilope25_bottom_heavy}e with a score of 0.8638. Denoising stabilizes this output, yielding smoother boundaries Figure~\ref{fig:antilope25_bottom_heavy}f, 0.9285 agreement), but still lacks the full spatial consistency of a refined model. When the threshold-guided supervision is applied, the image-only M-TGS refiner (Figure~\ref{fig:antilope25_bottom_heavy}g) cleanly connects the elongated features into internally consistent entities, achieving an excellent agreement of 0.9524. 

The multimodal M-TGS refiner shown in Figure~\ref{fig:antilope25_bottom_heavy}h is visually coherent and provides a slightly lower score of 0.9445, therefore, the learned refinement framework generalizes robustly across fundamentally different geometric structures. We also observe that when the acoustic image has a highly distinctive, high-contrast structural signal, multimodal concatenation provides no incremental benefit over image-only refinement.

\subsection{A localized anomaly in multimodal fusion}
\label{subsec:antilope25_middle_detailed}

\begin{figure}[htbp]
\centering
\includegraphics[width=.9\textwidth]{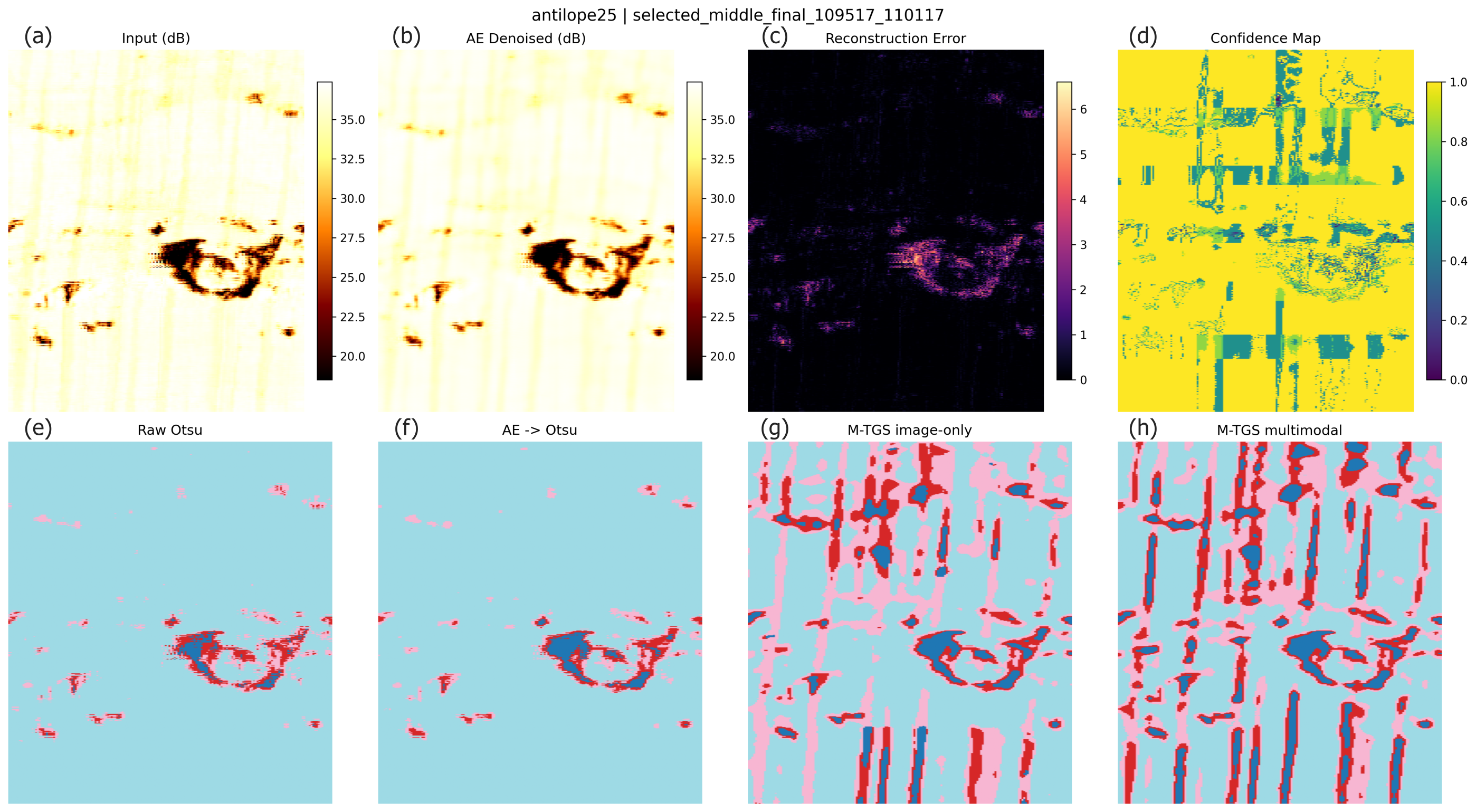}
\caption{Heavy-rerun interval from Antilope25 illustrating a challenging case for multimodal fusion. Top row: \textbf{(a)} input acoustic image, \textbf{(b)} autoencoder-denoised reconstruction, \textbf{(c)} reconstruction error, and \textbf{(d)} confidence map. Bottom row: \textbf{(e)} raw thresholding, \textbf{(f)} denoised thresholding, \textbf{(g)} image-only refined segmentation, and \textbf{(h)} multimodal refined segmentation (Concat). The interval contains a localized anomalous structure embedded within a weakly textured, vertically striped background. The image-only refiner successfully isolates the compact anomaly, whereas simple multimodal concatenation amplifies background noise.}
\label{fig:antilope25_middle_heavy}
\end{figure}

Figure~\ref{fig:antilope25_middle_heavy} illustrates the critical failure mode of simple multimodal concatenation using a highly challenging mid-depth interval from the Antilope25 well. Unlike the broad bands or continuous columns of the previous cases, the relevant structural signal here is a compact, localized anomaly embedded within a weakly textured, vertically striped background Figure~\ref{fig:antilope25_middle_heavy}a. The autoencoder preserves this principal target Figure~\ref{fig:antilope25_middle_heavy}b, with reconstruction errors appropriately highlighting both the core anomaly and structured variations in the background (Figure~\ref{fig:antilope25_middle_heavy}c). Figure~\ref{fig:antilope25_middle_heavy}d exhibits heterogeneous uncertainty that extends from the anomaly's boundaries into the vertical striping, signaling a highly complex provisional target.

Direct thresholding struggles severely with this compact geometry, recovering only a sparse, incomplete residual of the anomaly even after denoising (Figure~\ref{fig:antilope25_middle_heavy}e,f; agreement rising marginally from 0.6494 to 0.6577). In stark contrast, the image-only M-TGS refiner (Figure~\ref{fig:antilope25_middle_heavy}g) effectively leverages local 2D spatial context to consolidate the weak target, reorganizing the fragmented baseline into a continuous and coherent spatial object with a high score of 0.8105.

However, the unselective injection of 1D log data via the simple multimodal refiner Figure~\ref{fig:antilope25_middle_heavy}h) proves actively detrimental. Instead of improving the compact target, the multimodal model increases the amount of irrelevant vertical background striping, thereby redistributing the network's attention away from the localized anomaly. This geometric distraction causes the quantitative agreement to plummet to 0.5809, a performance very similar to the unsupervised latent-clustering, which has a score of 0.5782.  

When a target is geometrically compact, as in this anomaly, and the auxiliary channels correlate with background texture, simple channel-wise stacking biases the segmentation toward the background. The latter result indicates that early concatenation might require a more sophisticated integration mechanism capable of spatially selective, depth-aware cross-modal interaction.

\subsection{Benchmark synthesis of depth-aware multimodal refinement}
\label{subsec:phase5_benchmark_synthesis}

Our benchmark showed that threshold-type refinement outperforms classical references and revealed that simple multimodal concatenation can have inconsistent gains. When averaged across case studies, the image-only refiner obtained a mean agreement of $0.9346$ with the pseudo-label, compared to $0.8935$ for the concatenation-based multimodal refiner. This interval-dependent performance indicates that the limitation does not lie in the auxiliary logs themselves, but in their unselective fusion.

\begin{figure*}[t]
    \centering
    \includegraphics[width=.55\textwidth]{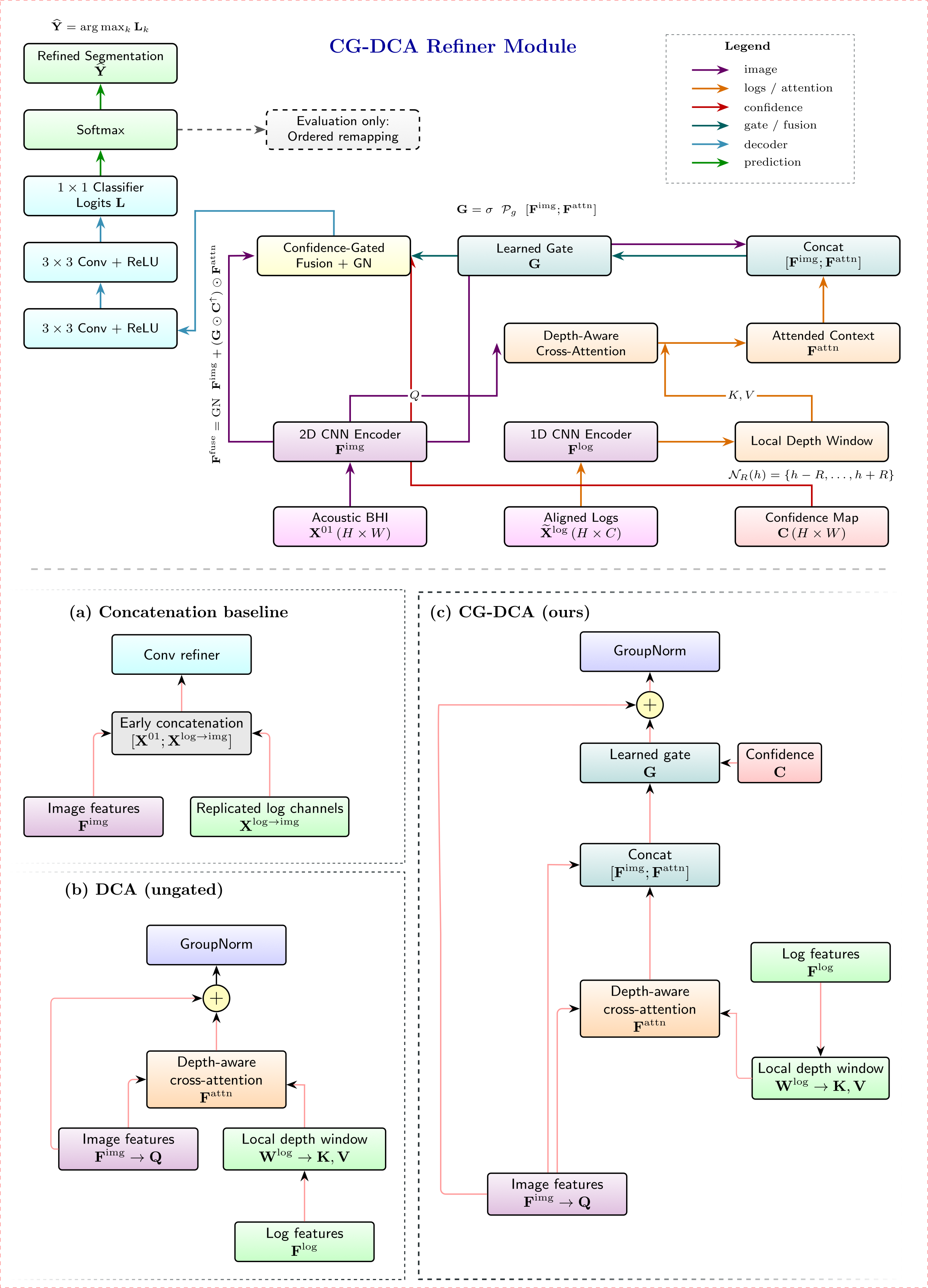}
    \caption{\textbf{Architectural evolution of the multimodal refinement framework.}
    \textbf{Top}, full schematic of the confidence-gated depth-aware cross-attention (CG-DCA) refiner, showing the image and log encoders, local depth window, cross-attention block, gating pathway, confidence-modulated fusion, decoder, and prediction stage.
    \textbf{Bottom}, conceptual comparison of the multimodal progression considered in this study: \textbf{(a)} early concatenation baseline, \textbf{(b)} ungated depth-aware cross-attention (DCA), and \textbf{(c)} confidence-gated DCA (CG-DCA).}
    \label{fig:figure10_architecture_evolution}
\end{figure*}

To address this, we evaluated a new family of multimodal refiners based on depth-aware cross-attention (Figure~\ref{fig:figure10_architecture_evolution}). Unlike simple channel concatenation, these models treat the image and logs as geometrically distinct but depth-aligned modalities. The baseline depth-aware cross-attention (DCA) model explicitly links each two-dimensional image row to one-dimensional conventional log measurements by allowing image features to query a local depth window of encoded log context. 

To further regulate this interaction, we introduced variants, including the proposed gated depth-aware cross-attention (G-DCA), which modulates the strength of the multimodal contribution via an explicit learned gate, and the confidence-gated depth-aware cross-attention (CG-DCA). Therefore, multimodal correction becomes not only depth-aware but spatially selective and reliability-aware. 

As detailed in Supplementary Table~S5, CG-DCA achieves the highest mean agreement against the pseudo-label reference across the 58 evaluation intervals ($0.8603$). The entire attention-based family improves performance, with G-DCA ($0.8201$) and DCA ($0.8098$) clearly outperforming both the earlier concatenation-based refiner ($0.7550$) and the image-only model ($0.7387$). The performance hierarchy is broadly stable across both the broad and deeper evaluation subsets and highly consistent at the interval level, with CG-DCA outperforming G-DCA in 55 of the 58 evaluated slices. 

These margins confirm that fusion is an important factor in multimodal performance. The unselective concatenation can dilute strong image-driven features. However, depth-aware, spatially selective, and confidence-modulated integration reliably converts auxiliary log context into systematic enhancements within segmentation coherence.


\subsubsection{Coala88 well case for multimodal fusion}
\label{subsec:coala88_heavy_top_detailed}
\begin{figure}[htbp]
\centering
\includegraphics[width=.9\textwidth]{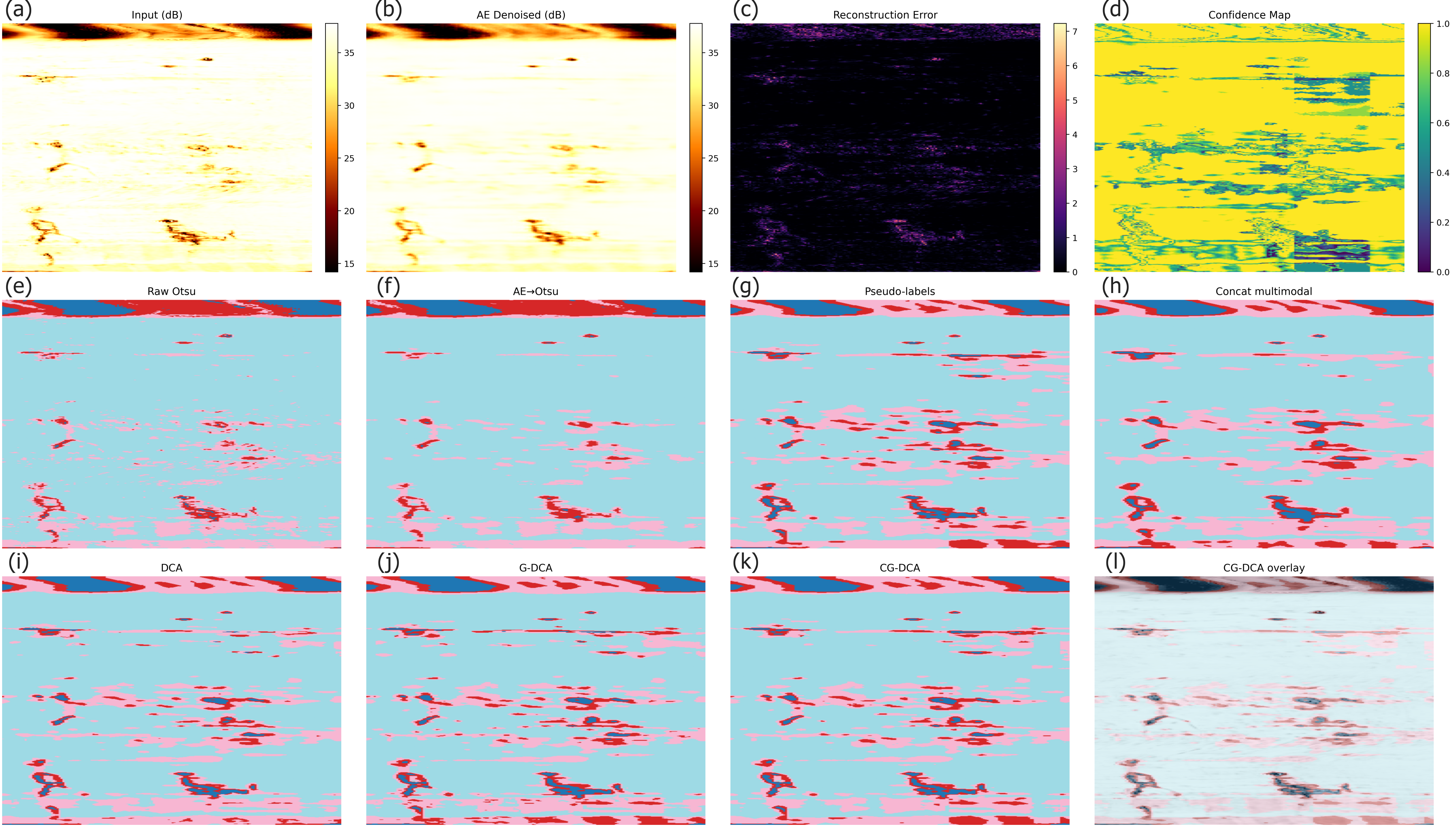}
\caption{Selected-case interval from Coala88 illustrating the progression from threshold-guided refinement to selectively fused multimodal refinement. Top row: \textbf{(a)} input acoustic image, \textbf{(b)} autoencoder-denoised reconstruction, \textbf{(c)} reconstruction error, and \textbf{(d)} confidence map. Middle row: \textbf{(e)} raw thresholding, \textbf{(f)} denoised thresholding, \textbf{(g)} pseudo-labels, and \textbf{(h)} concatenation-based multimodal refinement. Bottom row: \textbf{(i)} DCA, \textbf{(j)} G-DCA, \textbf{(k)} CG-DCA, and \textbf{(l)} CG-DCA overlay.}
\label{fig:coala88_heavy_top}
\end{figure}

Figure~\ref{fig:coala88_heavy_top} demonstrates the Phase 4 multimodal family's benefit in a visually subtle, although systematic case. The Coala88 slice features weak, spatially sparse structures embedded within a uniform bright background. It tests whether the fusion mechanisms improve segmentation selectively—by preserving faint elongated features—without creating spurious classes or over-segmenting the background.

The input image (Figure~\ref{fig:coala88_heavy_top}a) contains diffuse, low-contrast anomalies. The autoencoder keeps these weak targets while attenuating fine-scale noise (Figure~\ref{fig:coala88_heavy_top}b). The reconstruction error (Figure~\ref{fig:coala88_heavy_top}c) highlights where the denoiser departs most strongly from the input, while the confidence map (Figure~\ref{fig:coala88_heavy_top}d) identifies the regions of greatest structural ambiguity from threshold distance and local vote consistency. Together, these panels show that refinement should be concentrated in sparse, uncertain regions with feature-bearing sections, while leaving the highly confident background intact.

We observe that raw thresholding in Figure~\ref{fig:coala88_heavy_top}e fragments these sparse targets, and denoising provides only modest stabilization, as shown in Figure~\ref{fig:coala88_heavy_top}f. The pseudo-label map in Figure~\ref{fig:coala88_heavy_top}g establishes a more spatially organized provisional target, and the direct multimodal concatenation shown in Figure~\ref{fig:coala88_heavy_top}h propagates mid-interval responses indiscriminately, resulting in soft, blurred boundaries.

The depth-aware fundamentally resolves this diffusion. The DCA shown in Figure~\ref{fig:coala88_heavy_top}i better localizes the sparse anomalies, and the G-DCA shown in Figure~\ref{fig:coala88_heavy_top}j suppresses spurious background contrast. Our proposed CG-DCA, shown in Figure~\ref {fig:coala88_heavy_top}k, provides the most controlled map. The method preserves weak targets without diffuse class spread, as confirmed by its tight image overlay shown in Figure~\ref{fig:coala88_heavy_top}l. The optimal formulation is not the one that produces the most visually aggressive segmentation, but the one that maximizes structural continuity in ambiguous zones.

The Methods section provides the structural metrics that quantify this selectivity. The advanced attention models leave the vast majority of the segmentation intact, modifying only 7.13\% (DCA), 8.35\% (G-DCA), and 8.99\% (CG-DCA) of pixels relative to the concatenation baseline. Despite these limited changes, pseudo-label agreement improves substantially from 0.9014 (Concat) to 0.9739 (CG-DCA), accompanied by a reduction in off-diagonal confusion mass from 0.0986 to 0.0261. Because most of these spatial edits (66.3\% for CG-DCA) occur precisely within the low-confidence regions of the weak supervisory signal, the architecture acts as a highly targeted refinement mechanism rather than a global re-segmentation model.
\subsubsection{Botorosa47 well refiner family}
\label{subsec:botorosa47_phase4_detailed}
\begin{figure}[htbp]
\centering
\includegraphics[width=.9\textwidth]{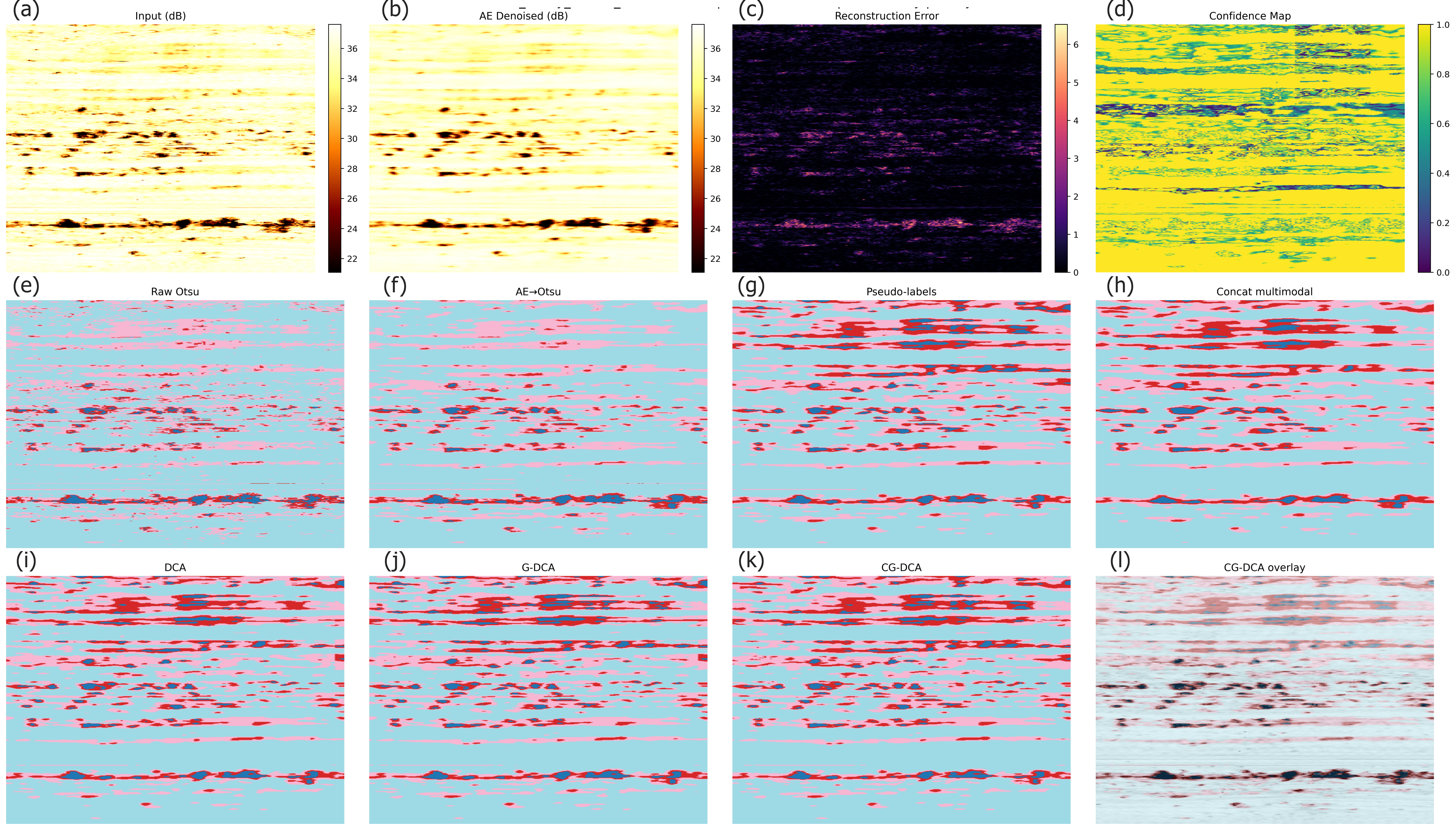}
\caption{Selected-case interval from Botorosa47 illustrating the progression from threshold-guided refinement to the Phase~4 multimodal family. Top row: \textbf{(a)} input acoustic image, \textbf{(b)} autoencoder-denoised reconstruction, \textbf{(c)} reconstruction error, and \textbf{(d)} confidence map. Middle row: \textbf{(e)} raw thresholding, \textbf{(f)} denoised thresholding, \textbf{(g)} pseudo-labels, and \textbf{(h)} concatenation-based multimodal refinement. Bottom row: \textbf{(i)} DCA, \textbf{(j)} G-DCA, \textbf{(k)} CG-DCA, and \textbf{(l)} CG-DCA overlay. In this interval, the main visual difference is between the earlier baselines and the DCA-family methods: the latter better preserve the laterally continuous banding visible in the image and reduce the diffuse class spread seen in the earlier multimodal formulation.}
\label{fig:botorosa47_phase4}
\end{figure}

Unlike the subtle improvements detected in the Coala88 interval, the Botorosa47 slice (Figure~\ref{fig:botorosa47_phase4}) presents a pronounced visual and structural contrast between the earlier unselective multimodal baseline and the depth-aware (DCA) family. This interval is dominated by laterally continuous horizontal banding, a central cluster of darker anomalies, and a distinct lower horizon (Figure~\ref{fig:botorosa47_phase4}a). The primary segmentation challenge is keeping this lateral continuity without introducing diffuse class spreading.

The autoencoder successfully prioritizes these horizontal structures (Figure~\ref{fig:botorosa47_phase4}b, c), allowing the pseudo-labels to form a highly coherent layered hypothesis (Figure~\ref{fig:botorosa47_phase4}g). However, the simple concatenation-based refiner (Figure~\ref{fig:botorosa47_phase4}h) fails to respect these crisp boundaries, smearing mid-interval responses heavily into adjacent vertical zones.

The introduction of depth-aware cross-attention fundamentally resolves this diffusion. The DCA family (Figure~\ref{fig:botorosa47_phase4}i--k) enforces strict spatial control, aligning the segmentation map with the laterally continuous architecture visible in the acoustic image. Among these variants, CG-DCA (Figure~\ref{fig:botorosa47_phase4}k, l) achieves the most precise structural recovery, balancing the preservation of thin banded features with the suppression of spurious localized noise.

The structural metrics from Section~\ref{subsec:structural_metrics_selected_cases} track this exact progression. Moving from unselective concatenation to baseline DCA corrects 9.08$\%$ of the pixel assignments, driving pseudo-label agreement up from 0.8806 to 0.9336. Adding learned gating (G-DCA) and confidence modulation (CG-DCA) pushes that agreement even higher to 0.9346 and 0.9568, respectively, which collapses the off-diagonal confusion mass to a mere 0.0432.

Furthermore, this architectural selectivity proves highly targeted. Of all the pixels modified by CG-DCA relative to G-DCA, 59.0$\%$ sit squarely inside the lowest-confidence quartile of the weak supervisory signal. The final performance leap of CG-DCA is therefore not a global smoothing artifact, but the result of surgically resolving geometrically complex horizons. It actively reorganizes the segmentation to match the dominant lateral banding of the raw acoustic data.

\subsubsection{Shallow Antilope25 as a case for family refinement}
\label{subsec:antilope25_phase4_top_detailed}
\begin{figure}[ht!]
\centering
\includegraphics[width=.9\textwidth]{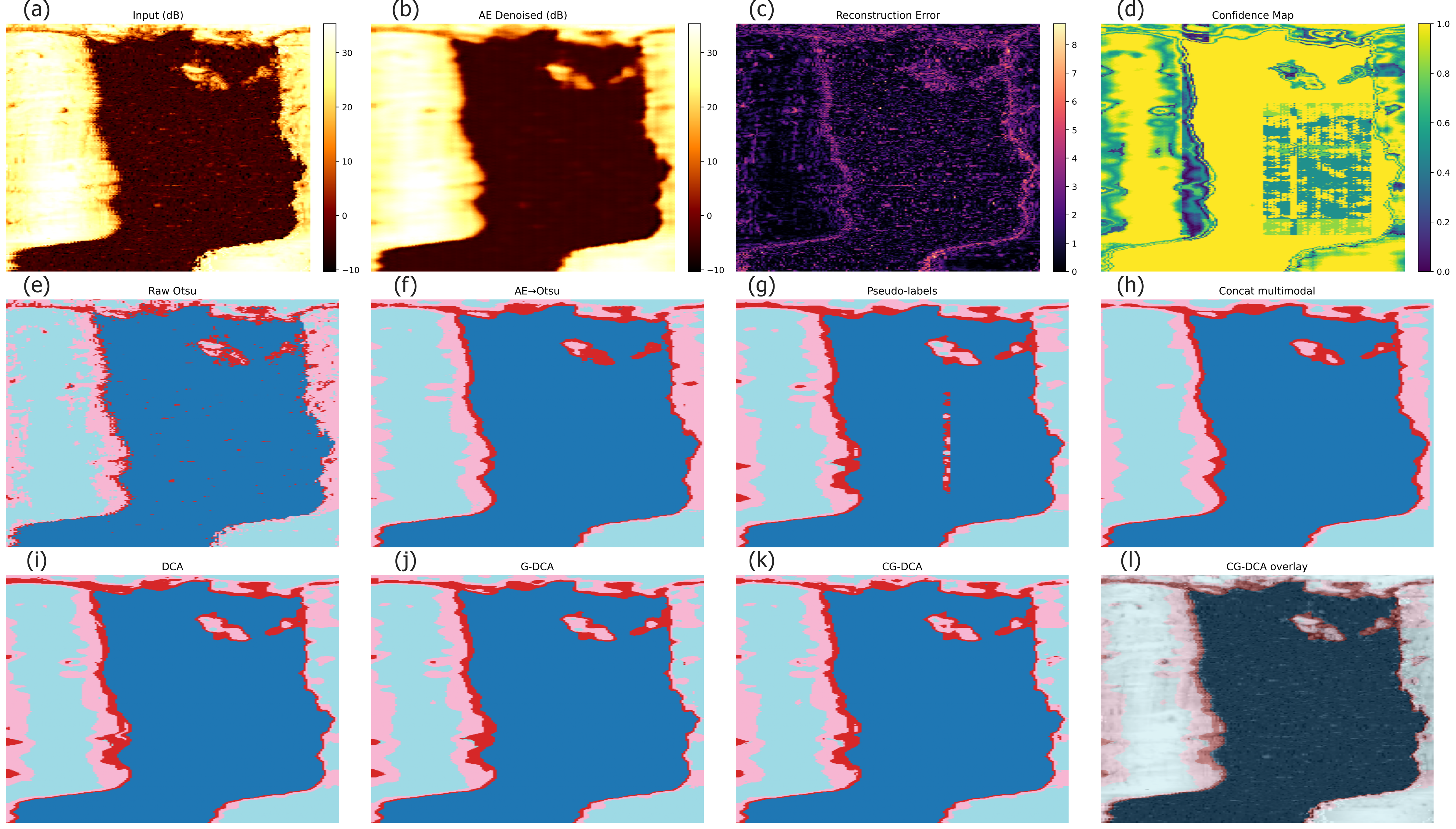}
\caption{Selected-case interval from Antilope25 illustrating the progression from threshold-guided refinement to the Phase~4 multimodal family. Top row: \textbf{(a)} input acoustic image, \textbf{(b)} autoencoder-denoised reconstruction, \textbf{(c)} reconstruction error, and \textbf{(d)} confidence map. Middle row: \textbf{(e)} raw thresholding, \textbf{(f)} denoised thresholding, \textbf{(g)} pseudo-labels, and \textbf{(h)} concatenation-based multimodal refinement. Bottom row: \textbf{(i)} DCA, \textbf{(j)} G-DCA, \textbf{(k)} CG-DCA, and \textbf{(l)} CG-DCA overlay. In this interval, the principal structure is a broad central dark body bounded by strong lateral transitions. Relative to the earlier baselines, the DCA-family methods recover this geometry more cleanly, and the progression from concat multimodal to DCA, G-DCA, and CG-DCA is visually more apparent than in the Botorosa47 case.}
\label{fig:antilope25_phase4_top}
\end{figure}

In contrast to the subtle textures of Coala88 and the fine banding of Botorosa47, the Antilope25 interval is dominated by a large central body (Figure~\ref{fig:antilope25_phase4_top}a).  This structure has sharp lateral boundaries and embedded localized anomalies. Here, the segmentation challenge shifts from recovering thin features to accurately outlining a dominant region. The goal is to avoid over-smoothing its margins or absorbing smaller internal components.

The autoencoder cleanly isolates this central body (Figure~\ref{fig:antilope25_phase4_top}b). The reconstruction error and confidence maps (Figure~\ref{fig:antilope25_phase4_top}c, d) correctly identify the sharp boundaries and upper anomalies as the main zones of ambiguity. The concatenation-based multimodal refiner (Figure~\ref{fig:antilope25_phase4_top}h) recovers the broad geometry, but operates bluntly. It smooths over localized responses and blurs detailed boundaries.

The transition to depth-aware fusion directly resolves these boundary ambiguities. The DCA family (Figure~\ref{fig:antilope25_phase4_top}i--k) provides stricter spatial control, detecting sharp transitions and isolating upper anomalies. The final overlay (Figure~\ref{fig:antilope25_phase4_top}l) confirms that the selective multimodal architecture preserves the dominant body and prevents class spread into the background.

Quantitative structural metrics (summarized for all three Phase 4 case studies in Supplementary Table~S6) reveal that this interval exhibits a strong but non-monotonic progression. The initial shift from concatenation to ungated DCA yields a massive performance leap, resolving 8.78\% of the pixel assignments and driving agreement from 0.8991 to 0.9702. The introduction of the learned gate (G-DCA) provides optimal refinement for this specific geometry, achieving 0.9765 agreement and reducing the off-diagonal confusion mass to just 0.0235. This gating step is highly targeted: 70.4\% of the pixels modified by G-DCA relative to DCA fall precisely within the lower-confidence regions of the supervisory signal.

While the confidence-gated variant (CG-DCA) remains highly accurate (0.9738), it actually yields a marginal regression compared to G-DCA in this specific slice. These three detailed case studies in Supplementary Table~S6 confirm the wider benchmark conclusions that depth-aware, selective fusion consistently surpasses unselective concatenation by actively targeting structurally ambiguous zones, even if the absolute optimal gating formulation remains mildly sensitive to local interval geometry.

\subsection{Targeted ablation and cross-well robustness}
\label{subsec:targeted_ablation_crosswell}
\begin{figure*}[htbp]
\centering
\includegraphics[width=.9\textwidth]{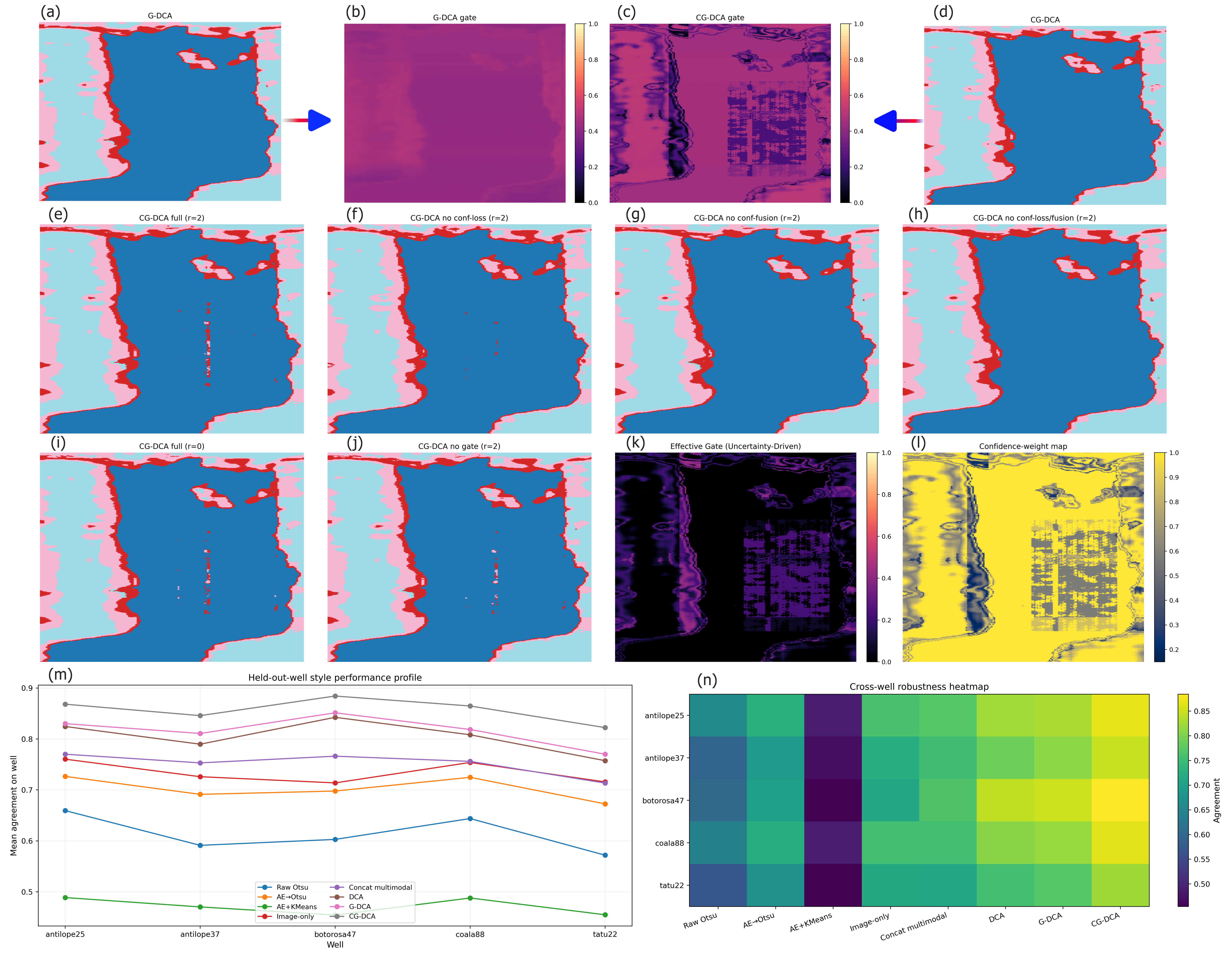}
\caption{\textbf{Targeted ablation and cross-well robustness of the final multimodal framework.}
\textbf{(a)} G-DCA segmentation. \textbf{(b)} G-DCA gate map. \textbf{(c)} CG-DCA gate map. \textbf{(d)} CG-DCA segmentation. Panels \textbf{(a)}--\textbf{(d)} illustrate how the confidence-gated variant produces a more structured and spatially selective multimodal control pattern than G-DCA in the representative Antilope25 heavy-top interval.
\textbf{(e)} Full CG-DCA with depth-window radius $r=2$. \textbf{(f)} CG-DCA without confidence-weighted loss. \textbf{(g)} CG-DCA without confidence-aware fusion. \textbf{(h)} CG-DCA without either confidence-weighted loss or confidence-aware fusion. \textbf{(i)} Full CG-DCA with $r=0$, that is, same-depth-only log interaction. \textbf{(j)} CG-DCA without the learned gate. \textbf{(k)} Effective uncertainty-driven fusion map. \textbf{(l)} Confidence-weight map used for pixelwise loss weighting. Together, panels \textbf{(e)}--\textbf{(l)} show that the strongest degradation arises when confidence-aware fusion is removed, whereas removing confidence-weighted loss alone or removing the gate has a smaller and more interval-dependent effect.
\textbf{(m)} Held-out-well style performance profile summarizing mean agreement on each well for all compared methods. \textbf{(n)} Cross-well robustness heatmap showing agreement against the pseudo-label reference across wells and methods.}
\label{fig:final_ablation_robustness}
\end{figure*}

We showed that selective depth-aware fusion improves benchmark-level agreement. To isolate the drivers of this gain and test its generalizability, we performed targeted ablations (Figure~\ref{fig:final_ablation_robustness}a--l) and cross-well robustness analyses (Figure~\ref{fig:final_ablation_robustness}m, n).

We observe that the learned multimodal control maps highlight the advantage of the confidence-gated architecture. The standard G-DCA model uses broad spatial modulation over the interval, as shown in Figure~\ref{fig:final_ablation_robustness}b. In contrast, the CG-DCA concentrates auxiliary-context fusion in ambiguous zones, like boundary transitions and irregular internal inclusions, as shown in Figure~\ref{fig:final_ablation_robustness}c. Therefore, confidence-aware fusion acts selectively rather than uniformly, targeting the harder regions of the segmentation problem.

We have investigated the quantitative ablation results, which support this interpretation, as shown in Table~\ref{tab:targeted_ablation_summary}. The full CG-DCA model with a depth-window radius of $r=2$ achieved a mean score of 0.9172 on the targeted ablation set. If we remove the confidence-aware fusion, we induce the largest degradation, reducing the score to 0.8904, whereas removing both confidence-aware mechanisms reduces it to 0.8914.  However, by contrast, removing only the confidence-weighted loss, ablating the explicit learned gate, or restricting attention to same-depth log interactions ($r=0$) did not consistently reduce performance on this specific subset, and all three variants remained close to the full model on average. In pairwise comparisons, the full model outperformed both variants lacking confidence-aware fusion across all 10 targeted intervals. These ablations isolate confidence-aware fusion, rather than confidence weighting alone, as the most critical contributor to the final model's behavior.

This structural advantage is robust across wells. Cross-well aggregation (Figure~\ref{fig:final_ablation_robustness}m, n) shows the same method hierarchy for all five wells. Therefore, thresholding references are weakest. The image-only and concatenation-based refinements are in the middle, and the DCA family is strongest. 

The well-wise profiles prove that this ranking is not driven by a single favorable well. CG-DCA remained the strongest method across all well-level aggregates. Furthermore, interval-level pairwise comparisons showed CG-DCA outperforming G-DCA in most slices of every well, including 21 of 22 intervals in Antilope25, 9 of 9 in Antilope37, 10 of 10 in Coala88, 8 of 8 in Tatu22, and 7 of 9 in Botorosa47. 

Therefore, the final performance gain is not a trivial consequence of increased complexity of the model, but rather depends specifically on confidence-aware, spatially selective multimodal interaction. Furthermore, the  advantage is not limited to a few favorable examples but remains stable under cross-well aggregation, firmly supporting the interpretation of CG-DCA as an effective refinement model within the present annotation-free benchmark.

\begin{table*}[htbp]
\centering
\caption{Cross-well robustness summary for all compared methods. Agreement is reported against the pseudo-label reference and therefore measures internal consistency within the annotation-free framework rather than externally validated geological accuracy. Mean over wells is the primary ranking criterion; the remaining columns summarize well-to-well variability.}
\label{tab:crosswell_main_methods}
\small
\begin{tabular}{lccccc}
\hline
Method & Mean over wells & Std.\ over wells & Min well & Max well & Range \\
\hline
Raw Otsu            & 0.6138 & 0.0366 & 0.5719 & 0.6594 & 0.0875 \\
AE$\rightarrow$Otsu & 0.7026 & 0.0230 & 0.6726 & 0.7265 & 0.0539 \\
AE+KMeans           & 0.4713 & 0.0168 & 0.4544 & 0.4887 & 0.0343 \\
Image-only          & 0.7339 & 0.0218 & 0.7138 & 0.7603 & 0.0466 \\
Concat multimodal   & 0.7518 & 0.0226 & 0.7134 & 0.7701 & 0.0567 \\
DCA                 & 0.8044 & 0.0328 & 0.7574 & 0.8425 & 0.0851 \\
G-DCA               & 0.8162 & 0.0299 & 0.7701 & 0.8514 & 0.0813 \\
CG-DCA              & \textbf{0.8571} & 0.0238 & 0.8223 & 0.8842 & 0.0620 \\
\hline
\end{tabular}
\end{table*}

\begin{table*}[htbp]
\centering
\caption{Targeted ablation summary for the CG-DCA family. Means are reported over the 10 targeted ablation intervals. The final columns summarize pairwise comparison against the full CG-DCA model with $r=2$ when such a comparison is defined. Negative $\Delta$ indicates lower agreement than the full model.}
\label{tab:targeted_ablation_summary}
\small
\begin{tabular}{lccccccc}
\hline
Variant & Overall mean & Std. & Broad mean & Heavy mean & $\Delta$ vs full $(r=2)$ & Full wins & Full win rate \\
\hline
CG-DCA full $(r=2)$                & 0.9172 & 0.0888 & 0.8783 & 0.9755 &  0.0000 & --    & --   \\
CG-DCA no conf-loss $(r=2)$        & 0.9187 & 0.0837 & 0.8842 & 0.9705 &  0.0016 & 3/10  & 0.30 \\
CG-DCA no conf-fusion $(r=2)$      & 0.8904 & 0.0984 & 0.8509 & 0.9497 & -0.0268 & 10/10 & 1.00 \\
CG-DCA no conf-loss/fusion $(r=2)$ & 0.8914 & 0.1006 & 0.8495 & 0.9543 & -0.0258 & 10/10 & 1.00 \\
CG-DCA full $(r=0)$                & 0.9181 & 0.0867 & 0.8792 & 0.9765 &  0.0010 & 6/10  & 0.60 \\
CG-DCA no gate $(r=2)$             & 0.9207 & 0.0861 & 0.8821 & 0.9786 &  0.0035 & 3/10  & 0.30 \\
\hline
\end{tabular}
\end{table*}

\section*{Discussion}

We have demonstrated that weakly supervised, threshold-based refinement can convert noisy, heuristic partitions of acoustic borehole images into coherent structural segmentations. Across the initial interval study, the compact cross-well screening, the deeper analyses, and the final attention-based benchmark, the most stable pattern is the substantial improvement gained by moving from raw or denoised thresholding to learned refinement. This result is reinforced by the consistently weak performance of the unsupervised AE + KMeans reference. The performance gain is therefore not driven by the mere use of learned features, but by the combination of denoising, pseudo-label generation, confidence estimation, and supervised-style refinement from a structured weak target.

The initial multimodal results showed that auxiliary logs are not always helpful when added by direct channel concatenation. In the compact screening, both the image-only and concatenation-type multimodal refiners achieved almost identical average scores. They also had a balanced interval-level win rate. In the laterally banded Botorosa47 interval, multimodal refinement clearly helped. However, in the vertically elongated Antilope25 interval, the image-only refiner was already strong and slightly outperformed concatenation. In a localized-anomaly Antilope25 interval, simple concatenation actually decreases performance. These results show that multimodal context can be valuable. Its usefulness depends on injecting the auxiliary signal in a way that fits the local image structure.

Deploying attention-based multimodal refiners substantially resolves this conflict. Once multimodal interaction is reformulated through depth-aware cross-attention and further regulated by learned gating and confidence-aware modulation, the benchmark-level ordering stabilizes. The confidence-gated depth-aware cross-attention (CG-DCA) model achieves the highest overall performance, and the family ranking remains broadly consistent across intervals and subsets, even though the strongest within-family variant can remain mildly interval-dependent. The latter suggests that possible limitations of multimodal fusion did not stem from the auxiliary logs but from the unselective way in which they were fused with the acoustic image. The acoustic image provides two-dimensional borehole-wall texture, whereas the logs provide a one-dimensional depth-indexed petrophysical context. A fusion method that respects this geometrical asymmetry, rather than treating replicated logs as ordinary image channels can be more effective.

We have observed that targeted ablation performance degrades most when confidence-aware fusion is removed. In contrast, removing only the confidence-weighted loss or limiting the network to pointwise depth interactions ($r=0$) does not always harm performance on the targeted subset. Both variants still show the role of uncertainty-aware and local depth-aware integration. The final model’s behavior is best explained by the interaction among local depth-aware attention, learned gating, and confidence-aware modulation, rather than by any component alone.

These findings are highly relevant because the absence of dense manual labels is a defining practical condition of borehole-image interpretation at scale. In such settings, a framework that begins with threshold-derived pseudo-labels and refines them through learned spatial regularization offers a highly practical middle ground between purely heuristic partitioning and fully supervised workflows. The present results show that this strategy is viable even before external geological validation is introduced. Therefore, we can generate internally consistent structural maps while preserving the  scalability of manual free-labelling analysis. Furthermore, the multimodal contributions and auxiliary logs specifically resolve existing image ambiguity rather than imposing structure indiscriminately.

Our research shows that threshold-guided learned refinement offers a strong path for segmenting geometrically complex borehole data under strict weak supervision. The multimodal fusion only attains its full potential when it is depth-aware, spatially selective, and confidence-modulated. We also observe that simple channel concatenation does not have sufficient precision to guarantee improvement. Furthermore, we conclude that the success of the DCA family suggests that a physically structured multimodal design is what actually converts auxiliary logs into a reliable source of segmentation value. Therefore, the CG-DCA architecture within the scope of this benchmark can be considered the most robust formulation for weakly supervised, multimodal borehole interpretation.

\section*{Methods}

\subsection{Problem formulation and multimodal geometry}

An acoustic borehole-image interval was represented as an unwrapped cylindrical wall image $\mathbf{X}^{\mathrm{img}}_{\mathrm{db}}\in\mathbb{R}^{H\times W}$, where $\mathbf{X}^{\mathrm{img}}_{\mathrm{db}}$ denotes the acoustic image in the decibel or amplitude domain, $H$ is the number of depth samples (rows), and $W$ is the number of azimuthal samples (columns) around the borehole wall. In parallel, depth-aligned conventional logs were represented as $\widetilde{\mathbf{X}}^{\log}\in\mathbb{R}^{H\times C}$ where $\widetilde{\mathbf{X}}^{\log}$ is the aligned log matrix, $H$ is the same depth dimension used by the image, and $C$ denotes the number of aligned auxiliary channels. In the final multimodal benchmark, $C=7$ channels were used: caliper (CAL), gamma ray (GR), bulk density (DEN), neutron porosity (NEU), compressional slowness (DTC), photoelectric factor (PE), and deep resistivity (RES90). We have considered an earlier interval-level analysis for experiment-specific channel subsets when one or more logs were unavailable or intentionally excluded. First, we have estimated a segmentation map, such that $\mathbf{Y}\in\{0,1,\dots,K-1\}^{H\times W}$ and $K=4$. The notation $\mathbf{Y}$ represents the predicted class map, each entry $Y(h,w)$ is an integer class label at depth row $h$ and azimuth column $w$, and $K=4$ is the total number of structural classes considered in the study. The classes were weakly supervised structural-amplitude classes induced by threshold-guided pseudo-labeling rather than manually annotated lithofacies. For interpretability, predicted classes were canonically reordered by increasing mean acoustic amplitude.

The benchmark was constructed from five wells (\textit{antilope25}, \textit{antilope37}, \textit{botorosa47}, \textit{coala88}, and \textit{tatu22}). The interval extraction yielded 69 intervals in total (54 broad and 15 representative heavy intervals) using slice height 600 rows, broad step 12{,}000 rows, minimum valid height 300 rows, and random seed 42. However, the central multimodal difficulty is geometric asymmetry, because the acoustic image resolves both depth and azimuth, whereas the auxiliary logs resolve only depth. Fusion must therefore inject depth-indexed context without fabricating azimuthal log structure.

\subsection{Preprocessing and depth alignment}

Missing image values were replaced by the interval median. For learning, the acoustic image in the amplitude domain was standardized as:
\begin{equation}
X^{z}(h,w)=\frac{X^{\mathrm{img}}_{\mathrm{db}}(h,w)-\mu}{\sigma},
\end{equation}
where $X^{z}(h,w)$ is the z-scored value at pixel $(h,w)$, $X^{\mathrm{img}}_{\mathrm{db}}(h,w)$ is the original amplitude at that location, $\mu$ is the mean of the interval, and $\sigma$ is the corresponding standard deviation. The standardized values were then softly compressed into $[0,1]$ through
\begin{equation}
X^{01}(h,w)=\frac{1}{2}\left[1+\tanh\!\left(\frac{X^{z}(h,w)}{3}\right)\right].
\end{equation}
Here, $X^{01}(h,w)$ denotes the normalized value used by the learning models, and the hyperbolic tangent compresses extreme values while preserving relative contrast around the center of the distribution. This produced the normalized image $\mathbf{X}^{01}\in[0,1]^{H\times W}$. In the final implementation, all learned refinement models used the denoised representation $\widehat{\mathbf{X}}^{01}$ rather than the raw normalized image.

Each conventional log channel was median-filled if necessary, robustly normalized by $1$--$99$ percentile clipping, and linearly interpolated to the image depth grid:
\begin{equation}
\widetilde{X}^{\log}(h,c)=\mathcal{I}_c\!\left(z_h^{\mathrm{img}}\right),
\qquad h=1,\dots,H,\;\; c=1,\dots,C,
\end{equation}
where $\widetilde{X}^{\log}(h,c)$ is the aligned value of channel $c$ at image depth row $h$, $\mathcal{I}_c(\cdot)$ denotes the interpolation operator for channel $c$, and $z_h^{\mathrm{img}}$ is the physical depth associated with image row $h$. For the direct concatenation baseline only, the aligned logs were laterally replicated across azimuth:
\begin{equation}
X^{\log\rightarrow\mathrm{img}}_c(h,w)=\widetilde{X}^{\log}(h,c),
\end{equation}
where $X^{\log\rightarrow\mathrm{img}}_c(h,w)$ denotes the image-shaped version replicated of the $c$-th log channel. This preserves depth synchrony, but does not assign azimuthal resolution to the logs.

\subsection{Interval-wise denoising autoencoder}

Pseudo-supervision was built from an interval-wise denoised image. Each normalized interval $\mathbf{X}^{01}$ was decomposed into overlapping $32\times32$ patches with stride $8$. A convolutional denoising autoencoder \cite{vincent2008extracting} encoded each patch through three stride-2 convolutional layers with channel progression $1\rightarrow16\rightarrow32\rightarrow64$, flattened the final feature map, and projected it onto a latent vector of dimension $d_z=64$. The decoder mirrored this topology by first expanding the latent vector through a fully connected layer and then reconstructing the patch with three transposed-convolution blocks $64\rightarrow32\rightarrow16\rightarrow1$ with sigmoid output. Gaussian noise with standard deviation $\sigma_\eta=0.05$ was injected during training.

Let $\mathbf{p}_i\in[0,1]^{32\times32}$ denote the $i$-th image patch extracted from the interval, where $i=1,\dots,N$ and $N$ is the total number of patches. Let $g_{\phi,\theta}$ denote the autoencoder, parameterized by encoder weights $\theta$ and decoder weights $\phi$. Training minimized the mean squared reconstruction error:
\begin{equation}
\mathcal{L}_{\mathrm{AE}}=
\frac{1}{N}\sum_{i=1}^{N}
\left\|
g_{\phi,\theta}(\mathbf{p}_i+\boldsymbol{\eta}_i)-\mathbf{p}_i
\right\|_2^2,
\qquad
\boldsymbol{\eta}_i\sim\mathcal{N}(0,0.05^2),
\end{equation}
where $\mathcal{L}_{\mathrm{AE}}$ is the autoencoder loss, $\boldsymbol{\eta}_i$ is additive Gaussian noise applied to patch $\mathbf{p}_i$, and $\|\cdot\|_2$ denotes the Euclidean norm over all patch pixels. Optimization used Adam \cite{kingma2014adam} with learning rate $10^{-3}$ and batch size 128. Training lasted 60 epochs for broad intervals and 120 epochs for heavy intervals.

Patch reconstructions were merged with a soft 2D Hann window with floor $\epsilon_{\mathrm{hann}}=0.05$ to obtain the denoised image $\widehat{\mathbf{X}}^{01}$. Here, $\epsilon_{\mathrm{hann}}$ prevents weights from vanishing at the patch boundaries. This denoised image was inverse-transformed back to the acoustic amplitude domain:
\begin{equation}
\widehat{\mathbf{X}}^{\mathrm{img}}_{\mathrm{db}}
=
3\sigma\,\operatorname{arctanh}(2\widehat{\mathbf{X}}^{01}-1)+\mu,
\end{equation}
where $\widehat{\mathbf{X}}^{\mathrm{img}}_{\mathrm{db}}\in\mathbb{R}^{H\times W}$ is the denoised image in the original acoustic-amplitude domain, $\mu$ and $\sigma$ are the mean and standard deviation used in the forward normalization of the interval, and $\operatorname{arctanh}(\cdot)$ is the inverse hyperbolic tangent, which reverses the earlier $\tanh$-based compression. The same patch-level latent vectors were retained for the latent-clustering baseline. For interval-level diagnostics and figure panels, a reconstruction-error map was also computed in the acoustic-amplitude domain:
\begin{equation}
E_{\mathrm{db}^2}(h,w)=\left(X^{\mathrm{img}}_{\mathrm{db}}(h,w)-\widehat{X}^{\mathrm{img}}_{\mathrm{db}}(h,w)\right)^2,
\end{equation}
where $E_{\mathrm{db}^2}(h,w)$ is the squared reconstruction error at pixel $(h,w)$, $X^{\mathrm{img}}_{\mathrm{db}}(h,w)$ is the original acoustic-amplitude value at that location, and $\widehat{X}^{\mathrm{img}}_{\mathrm{db}}(h,w)$ is the corresponding denoised reconstruction. Together with its log-compressed version, as shown by the following equation:
\begin{equation}
E^{\log}(h,w)=\log\!\left(1+E_{\mathrm{db}^2}(h,w)\right).
\end{equation}
where $E^{\log}(h,w)$ is the log-compressed reconstruction error, and the term $1+E_{\mathrm{db}^2}(h,w)$ ensures numerical stability when the squared error is zero. These maps were used for visualization and selected-case analysis, but not as direct inputs to the final confidence map used in the benchmark.

\subsection{Weak supervision: pseudo-labels and confidence}

Weak supervision was derived from the denoised image $\widehat{\mathbf{X}}^{\mathrm{img}}_{\mathrm{db}}$. First, global four-class Multi-Otsu thresholding \cite{otsu1979threshold,liao2001} produced
\begin{equation}
\mathbf{Y}^{\mathrm{AE\rightarrow Otsu}}
=
\operatorname{MO}\!\left(\widehat{\mathbf{X}}^{\mathrm{img}}_{\mathrm{db}},K=4\right),
\end{equation}
where $\mathbf{Y}^{\mathrm{AE\rightarrow Otsu}}$ is the global threshold-based segmentation of the denoised image and $\operatorname{MO}(\cdot,K)$ denotes the Multi-Otsu operator of class $K$. Second, an adaptive local Multi-Otsu procedure\cite{otsu1979threshold,liao2001} was applied on overlapping windows of size $128\times64$ with overlap $32\times16$. The tile votes were aggregated into a pixelwise label field, and the winning label was smoothed with a $3\times3$ median filter\cite{huang1979fast} to obtain the pseudo-label map:

\begin{equation}
Y^{\mathrm{pseudo}}(h,w)
=
\operatorname{median}
\left\{
Y^{\mathrm{local}}(i,j)\;:\; (i,j)\in\mathcal{N}_{3\times3}(h,w)
\right\}.
\end{equation}
Here, $\mathbf{Y}^{\mathrm{local}}$ is the label map obtained after aggregating the local tile predictions, and $\mathcal{N}_{3\times3}(h,w)$ denotes the $3\times3$ neighborhood centered at pixel $(h,w)$. The final pseudo-label map was therefore the locally regularized threshold-derived segmentation of the denoised image, whereas global AE$\rightarrow$Otsu was retained as a separate baseline.

A confidence map $\mathbf{C}\in[0,1]^{H\times W}$ quantified heuristic reliability by combining a global threshold-distance term and a local vote-margin term. If $\{\tau_m\}$ are the global Multi-Otsu thresholds, such that:
\begin{equation}
C_{\mathrm{global}}(h,w)
=
\min\!\left(
1,\,
\max\!\left(
0,\,
\frac{\min_m\left|\widehat{X}^{\mathrm{img}}_{\mathrm{db}}(h,w)-\tau_m\right|}{0.25\,\mathrm{sd}(\widehat{\mathbf{X}}^{\mathrm{img}}_{\mathrm{db}})+\varepsilon}
\right)
\right).
\end{equation}
where $C_{\mathrm{global}}(h,w)$ is the global confidence term at pixel $(h,w)$, $\tau_m$ are the global threshold values, $\mathrm{sd}(\cdot)$ denotes standard deviation, and $\varepsilon$ is a small numerical stabilizer preventing division by zero. If $v_{(1)}(h,w)$ and $v_{(2)}(h,w)$ are the largest and second-largest local vote counts at pixel $(h,w)$, then we have the following equation:
\begin{equation}
C_{\mathrm{local}}(h,w)
=
\frac{v_{(1)}(h,w)-v_{(2)}(h,w)}{v_{(1)}(h,w)+\varepsilon}.
\end{equation}
where $C_{\mathrm{local}}(h,w)$ measures the dominance of the winning local label vote over the runner-up vote. The final confidence map was obtained by averaging the global and local confidence terms and restricting the result to the interval $[0,1]$:
\begin{equation}
C(h,w)=
\min\!\left(
1,\,
\max\!\left(
0,\,
0.5\,C_{\mathrm{global}}(h,w)+0.5\,C_{\mathrm{local}}(h,w)
\right)
\right).
\end{equation}
where equal weights were given to the global and local confidence components. For confidence-weighted optimization, the corresponding weight map can computed with the equation:
\begin{equation}
W^{\mathrm{conf}}(h,w)=\lambda+(1-\lambda)C(h,w)^\gamma,
\end{equation}
where $W^{\mathrm{conf}}(h,w)$ is the pixel-wise optimization weight, $\lambda$ is a confidence floor, and $\gamma$ controls the nonlinearity of the confidence weighting. In the final confidence-gated model, $\lambda=0.15$ and $\gamma=1.0$. For qualitative interval figures, the standard visualization sequence comprised the input image $\mathbf{X}^{\mathrm{img}}_{\mathrm{db}}$, the denoised reconstruction $\widehat{\mathbf{X}}^{\mathrm{img}}_{\mathrm{db}}$, the reconstruction-error map $E^{\log}$, the confidence map $\mathbf{C}$, threshold baselines, pseudo-labels, and the corresponding learned predictions.

\subsection{Baseline methods}

The thresholding baselines were direct four-class quantizations of the raw and denoised images:
\begin{equation}
\mathbf{Y}^{\mathrm{RawOtsu}}
=
\mathcal{Q}_{\mathcal{T}^{\mathrm{raw}}}\!\left(\mathbf{X}^{\mathrm{img}}_{\mathrm{db}}\right),
\qquad
\mathbf{Y}^{\mathrm{AE\rightarrow Otsu}}
=
\mathcal{Q}_{\mathcal{T}^{\mathrm{den}}}\!\left(\widehat{\mathbf{X}}^{\mathrm{img}}_{\mathrm{db}}\right),
\end{equation}
where $\mathcal{Q}_{\mathcal{T}}(\cdot)$ denotes quantization under threshold set $\mathcal{T}$, $\mathcal{T}^{\mathrm{raw}}$ are the thresholds fitted to the raw image, and $\mathcal{T}^{\mathrm{den}}$ are those fitted to the denoised image.

A latent-clustering baseline was obtained by applying $K$-means\cite{mcqueen1967some} with $K=4$, $n_{\mathrm{init}}=10$, and random state 42 to standardized patch descriptors formed by concatenating the autoencoder latent vector with the mean patch amplitude in the original acoustic domain. Patch assignments were projected back to pixels with the same soft Hann aggregation used for reconstruction.

Two learned refinement baselines were then considered. The image-only refiner used $\mathbf{U}^{\mathrm{img}}=\widehat{\mathbf{X}}^{01}$ where $\mathbf{U}^{\mathrm{img}}$ is the input tensor supplied to the image-only refinement network. The direct multimodal concatenation baseline used is represented by the equation:
\begin{equation}
\mathbf{U}^{\mathrm{mm}}=
\mathrm{concat}\!\left(\widehat{\mathbf{X}}^{01},\mathbf{L}_1,\dots,\mathbf{L}_C\right),
\end{equation}
where $\mathbf{U}^{\mathrm{mm}}$ is the multimodal input tensor and $\mathbf{L}_c$ is the laterally replicated image-shaped version of the $c$-th aligned log channel, with entries $L_{c,h,w}=\widetilde{X}^{\log}(h,c)$. Both used the same shallow convolutional refiner with channel progression $d_{\mathrm{in}}\rightarrow32\rightarrow64\rightarrow64\rightarrow4$
where $d_{\mathrm{in}}=1$ for the image-only baseline and $d_{\mathrm{in}}=1+C$ for the direct multimodal baseline. The network was implemented with $3\times3$ convolutions and ReLU activations\cite{nair2010relu}, followed by a final $1\times1$ classifier. These models were optimized with Adam at learning rate $10^{-3}$ for 80 epochs on broad intervals and 140 epochs on heavy intervals, using standard pixelwise cross-entropy.

\subsection{Depth-aware cross-attention refinement}

As summarized in Figure~\ref{figure10_architecture_evolution}, the multimodal architecture comprised a 2D image encoder, a 1D log encoder, a local depth-window attention block, a learned gate, confidence-gated fusion, a shallow classifier head, and an evaluation-only ordered-remapping stage. To respect modality geometry, the denoised image and aligned logs were encoded separately. The image encoder was a 2D convolutional network with channel progression $1\rightarrow32\rightarrow64\rightarrow64$, and the log encoder was a 1D convolutional network with channel progression $C\rightarrow32\rightarrow64$. This produced the following attributes:
\begin{equation}
\mathbf{F}^{\mathrm{img}}=
g_{\theta_{\mathrm{img}}}(\widehat{\mathbf{X}}^{01})
\in\mathbb{R}^{D\times H\times W},
\qquad
\mathbf{F}^{\log}=
g_{\theta_{\log}}(\widetilde{\mathbf{X}}^{\log\top})
\in\mathbb{R}^{D\times H},
\end{equation}
where $\mathbf{F}^{\mathrm{img}}$ is the encoded image feature tensor, $\mathbf{F}^{\log}$ is the encoded log feature tensor, $g_{\theta_{\mathrm{img}}}$ and $g_{\theta_{\log}}$ are the image and log encoders, and $D=64$ is the feature dimension. In all depth-aware cross-attention (DCA) family models, four attention heads were used and the depth-window radius was $r=2$, corresponding to a local neighborhood of $2r+1=5$ depth positions.

For each image row $h$, the azimuthal image features formed query tokens $\mathbf{Q}_h\in\mathbb{R}^{W\times D}$, where $h\in\{1,\dots,H\}$ indexes the depth row, $W$ is the number of azimuthal positions in that row, and $D$ is the feature dimension of the encoded representation. Thus, each row $h$ is represented by $W$ query vectors of dimension $D$, one for each azimuthal location. The log features within the local depth neighborhood
\begin{equation}
\mathcal{N}(h;r)=\{\max(1,h-r),\dots,\min(H,h+r)\}
\end{equation}
were used to construct the keys and values. Here, $\mathcal{N}(h;r)$ denotes the set of depth indices within radius $r$ of row $h$, clipped to the valid interval range $\{1,\dots,H\}$. Therefore, the attention mechanism does not use the entire log column at once, but only a local vertical window centered on the current image row. The attended log context for row $h$ was then computed as:
\begin{equation}
\mathbf{A}_h=
\operatorname{MHA}\!\left(
\mathbf{Q}_hW_Q,\mathbf{K}_hW_K,\mathbf{V}_hW_V
\right),
\end{equation}
where $\mathbf{A}_h$ is the attended context tensor for image row $h$, $\mathbf{K}_h$ and $\mathbf{V}_h$ are the key and value tensors formed from the encoded log features in the neighborhood $\mathcal{N}(h;r)$, $W_Q$, $W_K$, and $W_V$ are the learned query, key, and value projection matrices, and $\operatorname{MHA}(\cdot)$ denotes multi-head attention\cite{vaswani2017attention}. In other words, $\mathbf{A}_h$ is the log-derived contextual representation retrieved for the $h$-th image row after matching the image queries to the depth-local log keys and values.

Equivalently, within each head the attention weights were given by the following equation:
\begin{equation}
\alpha_{h,w,r}
=
\frac{
\exp\!\left(\mathbf{q}_{h,w}^{\top}\mathbf{k}_{r}/\sqrt{D_h}\right)
}{
\sum_{r' \in \mathcal{N}(h;r)}
\exp\!\left(\mathbf{q}_{h,w}^{\top}\mathbf{k}_{r'}/\sqrt{D_h}\right)
},
\end{equation}
where $\alpha_{h,w,r}$ is the attention weight assigned by image location $(h,w)$ to the $r$-th log position in the local depth neighborhood, $\mathbf{q}_{h,w}$ is the query vector at that image location, $\mathbf{k}_r$ is the corresponding key vector, and $D_h=D/4$ is the per-head feature dimension. The attended log context at location $(h,w)$ was then computed with the equation:
\begin{equation}
\mathbf{a}_{h,w}
=
\sum_{r\in\mathcal{N}(h;r)} \alpha_{h,w,r}\mathbf{v}_r,
\end{equation}
where $\mathbf{v}_r$ is the value vector at the $r$-th depth position in the local neighborhood. In the basic DCA model, the attended log context was injected through a residual Layer Normalization block:
\begin{equation}
\widetilde{\mathbf{F}}^{\mathrm{DCA}}_h=
\operatorname{LN}\!\left(\mathbf{Q}_hW_Q+\mathbf{A}_h\right),
\end{equation}
where $\widetilde{\mathbf{F}}^{\mathrm{DCA}}_h$ is the fused representation at row $h$ after residual attention and $\operatorname{LN}(\cdot)$ denotes Layer Normalization\cite{ba2016layer}. The full DCA representation was obtained by stacking the row-wise outputs over depth:
\begin{equation}
\mathbf{F}^{\mathrm{DCA}}=\operatorname{stack}_{h}\!\left(\widetilde{\mathbf{F}}^{\mathrm{DCA}}_h\right).
\end{equation}
where  $\operatorname{stack}_h(\cdot)$ denotes depth-wise stacking of the row-level outputs to reconstruct a full 2D feature tensor. The gated depth-aware cross-attention (G-DCA) variant introduced a learned spatial gate, as shown in the following equation:
\begin{equation}
\mathbf{G}=
\sigma\!\left(
\phi\bigl([\mathbf{F}^{\mathrm{img}};\mathbf{F}^{\mathrm{DCA}}]\bigr)
\right),
\end{equation}
where $\mathbf{G}\in[0,1]^{D\times H\times W}$ is the learned gate tensor, $[\cdot;\cdot]$ denotes channel-wise concatenation, $\phi$ is a two-layer $1\times1$ convolutional network with channel progression $2D\rightarrow D\rightarrow D$ and a ReLU nonlinearity\cite{nair2010relu} after the first layer, and $\sigma$ is the sigmoid. The gate therefore learned, at each depth, azimuth, and feature channel, how strongly the log-conditioned correction should modulate the image representation. Therefore, the definition of Fusion in the following equation:
\begin{equation}
\mathbf{Z}^{\mathrm{G}}=
\operatorname{GN}\!\left(
\mathbf{F}^{\mathrm{img}}+\mathbf{G}\odot\mathbf{F}^{\mathrm{DCA}}
\right),
\end{equation}
where $\mathbf{Z}^{\mathrm{G}}$ is the gated fused representation, $\odot$ denotes element-wise multiplication, and $\operatorname{GN}(\cdot)$ denotes Group Normalization\cite{wu2018groupnorm} with 8 groups. The confidence-gated depth-aware cross-attention (CG-DCA) further modulated the multimodal correction by the confidence map broadcast along the channel dimension:
\begin{equation}
\bar{\mathbf{C}}\in[0,1]^{D\times H\times W},
\qquad
\bar{C}_{d,h,w}=C(h,w),
\end{equation}
and the fused representation can be written as the following equation:
\begin{equation}
\mathbf{Z}^{\mathrm{CG}}=
\operatorname{GN}\!\left(
\mathbf{F}^{\mathrm{img}}+
(\mathbf{G}\odot\bar{\mathbf{C}})\odot\mathbf{F}^{\mathrm{DCA}}
\right),
\end{equation}
where $\mathbf{Z}^{\mathrm{CG}}$ is the confidence-gated fused representation. This formulation corresponds to the confidence-gated fusion block shown in Figure~\ref{fig:figure10_architecture_evolution}: multimodal correction was attenuated both where the learned gate suppressed it and where pseudo-label confidence was low. A shallow 2D classifier head with channel progression $64\rightarrow64\rightarrow32\rightarrow4$ produced class logits:
\begin{equation}
\mathbf{S}^{m}=d_{\theta_{\mathrm{cls}}}(\mathbf{Z}^{m}),
\qquad
m\in\{\mathrm{DCA},\mathrm{G},\mathrm{CG}\},
\end{equation}
where $\mathbf{S}^{m}\in\mathbb{R}^{K\times H\times W}$ is the logit tensor for model variant $m$, $d_{\theta_{\mathrm{cls}}}$ is the classifier head, and $\mathbf{Z}^{m}$ denotes the fused representation for the corresponding model. Class probabilities were obtained by the following equation:
\begin{equation}
p_k(h,w)=\frac{\exp(S^m_{k,h,w})}{\sum_{j=0}^{K-1}\exp(S^m_{j,h,w})},
\end{equation}
and predictions were then obtained as
\begin{equation}
\widehat{Y}^{m}_{h,w}=\arg\max_k p_k(h,w),
\end{equation}
where $\widehat{Y}^{m}_{h,w}$ is the predicted class at pixel $(h,w)$. The ordered remapping stage shown in Figure 11 was applied only for evaluation and visualization, and was not part of model training.

\subsection{Optimization and targeted ablations}

All learned refinement models were trained against $\mathbf{Y}^{\mathrm{pseudo}}$. The image-only, direct-concatenation, DCA, and G-DCA models used standard pixelwise cross-entropy, as defined in the following equation:
\begin{equation}
\mathcal{L}_{\mathrm{CE}}
=
-\frac{1}{HW}\sum_{h,w}\log p_{Y^{\mathrm{pseudo}}(h,w)}(h,w),
\end{equation}
where $\mathcal{L}_{\mathrm{CE}}$ is the average pixelwise cross-entropy and $p_{Y^{\mathrm{pseudo}}(h,w)}(h,w)$ is the predicted probability assigned to the pseudo-label at pixel $(h,w)$. CG-DCA used confidence-weighted cross-entropy defined in the following equation:
\begin{equation}
\mathcal{L}_{\mathrm{CG\mbox{-}DCA}}
=
-\frac{1}{HW}\sum_{h,w}W^{\mathrm{conf}}(h,w)\log p_{Y^{\mathrm{pseudo}}(h,w)}(h,w).
\end{equation}
Here, $\mathcal{L}_{\mathrm{CG\mbox{-}DCA}}$ differs from $\mathcal{L}_{\mathrm{CE}}$ only by the inclusion of the pixel-wise confidence weight $W^{\mathrm{conf}}(h,w)$.

The DCA-family benchmark models were optimized with Adam at learning rate $10^{-3}$ for 180 epochs on broad intervals and 1000 epochs on heavy intervals. Training was performed on complete intervals rather than mini-batches of crops. Targeted ablations were then carried out on 10 selected intervals (six best-performing and four difficult cases, with best-per-well inclusion) using the same 180/1000 training schedule. The tested variants were: full CG-DCA with $r=2$; no confidence-weighted loss; no confidence-gated fusion; neither confidence mechanism; full CG-DCA with $r=0$; and no learned gate. Here, setting $r=0$ removes the local depth neighborhood and reduces cross-attention to a degenerate single-depth interaction, while removing the gate eliminates the learned spatial modulation of multimodal fusion. These experiments isolated the effects of local depth context, learned gating, and confidence use in fusion and optimization.

\subsection{Permutation-invariant evaluation}

Because weakly supervised class identities are permutation-ambiguous, predicted classes from clustering and learned models were reordered by their mean acoustic amplitude in $\mathbf{X}^{\mathrm{img}}_{\mathrm{db}}$. For a segmentation $\widehat{\mathbf{Y}}$, class $k$ was assigned mean amplitude:
\begin{equation}
\widehat{\mu}_k
=
\frac{1}{|\widehat{\mathcal{C}}_k|}
\sum_{(h,w)\in\widehat{\mathcal{C}}_k}
X^{\mathrm{img}}_{\mathrm{db}}(h,w),
\end{equation}
where $\widehat{\mu}_k$ is the mean acoustic amplitude of class $k$ and $\widehat{\mathcal{C}}_k=\{(h,w):\widehat{Y}_{h,w}=k\}$ is the set of pixels assigned to that class. Classes were then relabeled in ascending order of $\widehat{\mu}_k$. Threshold-based maps are intrinsically ordered by their thresholds and therefore already satisfy this convention. This ordering was used for interpretability, whereas evaluation robustness was ensured separately through optimal permutation matching.

For two segmentations $\mathbf{Y}^{A}$ and $\mathbf{Y}^{B}$, the contingency matrix is represented by the following equation:
\begin{equation}
M_{i,j}=\#\{(h,w):Y^{A}_{h,w}=i,\;Y^{B}_{h,w}=j\},
\end{equation}
where $M_{i,j}$ counts the number of pixels jointly assigned to class $i$ in $\mathbf{Y}^{A}$ and class $j$ in $\mathbf{Y}^{B}$. The optimal permutation was obtained by Hungarian matching\cite{kuhn1955,munkres1957}, such that:
\begin{equation}
\pi^{\star}
=
\arg\max_{\pi\in\mathfrak{S}_4}\sum_{i=0}^{3}M_{i,\pi(i)},
\end{equation}
where $\mathfrak{S}_4$ is the set of all permutations of the four class labels and $\pi^{\star}$ is the permutation that maximizes matched contingency mass. The permutation-invariant agreement was then computed with the following equation:
\begin{equation}
\mathrm{Acc}_{\mathrm{perm}}(\mathbf{Y}^{A},\mathbf{Y}^{B})
=
\frac{1}{HW}\sum_{h,w}\mathbb{1}\!\left[Y^{A}_{h,w}=\pi^{\star}(Y^{B}_{h,w})\right].
\end{equation}
Here, $\mathbb{1}[\cdot]$ is the indicator function, equal to 1 when its argument is true and 0 otherwise. This was the principal metric used against the pseudo-label reference in all compact benchmark comparisons. Agreement against AE$\rightarrow$Otsu and raw Otsu, aligned confusion matrices, class fractions, and per-well summaries were also recorded.

For selected-case structural analysis, all compared segmentations were first aligned to the pseudo-label reference. If $\mathbf{M}$ denotes the aligned confusion matrix, the diagonal and off-diagonal confusion masses were defined as
\begin{equation}
\operatorname{DiagMass}(\mathbf{M})=\frac{\operatorname{tr}(\mathbf{M})}{\sum_{i,j}M_{i,j}},
\qquad
\operatorname{OffDiagMass}(\mathbf{M})=1-\operatorname{DiagMass}(\mathbf{M}).
\end{equation}
where $\operatorname{tr}(\mathbf{M})=\sum_i M_{i,i}$ is the trace of $\mathbf{M}$, that is, the total mass on the diagonal, and $\sum_{i,j}M_{i,j}$ is the total number of evaluated pixels. Therefore, $\operatorname{DiagMass}(\mathbf{M})$ measures the fraction of aligned pixel assignments lying on the diagonal of the confusion matrix, whereas $\operatorname{OffDiagMass}(\mathbf{M})$ measures the fraction lying off the diagonal. Relative to the direct-concatenation baseline $\mathbf{Y}^{\mathrm{concat}}$, the fraction of changed pixels for method $m$ was
\begin{equation}
\operatorname{ChangedFrac}(m)=
\frac{1}{HW}\sum_{h,w}\mathbb{1}\!\left[Y^{m}(h,w)\neq Y^{\mathrm{concat}}(h,w)\right].
\end{equation}
Here, $\mathbf{Y}^{m}$ denotes the aligned segmentation produced by method $m$, $H$ and $W$ are the depth and azimuthal dimensions of the interval, and $\mathbb{1}[\cdot]$ is the indicator function, which is equal to $1$ when the condition inside the brackets is true and $0$ otherwise. Thus, $\operatorname{ChangedFrac}(m)$ measures the fraction of pixels whose class assignment differs between method $m$ and the concatenation baseline. Low-confidence regions were defined in quantile mode by the first quartile of the confidence map,
\begin{equation}
\tau_{\mathrm{low}}=Q_{0.25}(\mathbf{C}),
\end{equation}
where $\mathbf{C}\in[0,1]^{H\times W}$ is the pixel-wise confidence map and $Q_{0.25}(\mathbf{C})$ denotes its 25th percentile. In other words, $\tau_{\mathrm{low}}$ is the threshold below which the lowest-confidence quarter of pixels lies. The fraction of changed pixels falling in low-confidence regions was defined as:
\begin{equation}
\operatorname{LowConfChangeFrac}(m)=
\frac{
\sum_{h,w}
\mathbb{1}\!\left[Y^{m}(h,w)\neq Y^{\mathrm{concat}}(h,w)\right]
\mathbb{1}\!\left[C(h,w)<\tau_{\mathrm{low}}\right]
}{
\sum_{h,w}\mathbb{1}\!\left[Y^{m}(h,w)\neq Y^{\mathrm{concat}}(h,w)\right]
}.
\end{equation}
This quantity therefore measures the proportion of all pixels changed by method $m$, relative to the concatenation baseline, that fall inside the least confident part of the interval. The boundary complexity was quantified by the total number of adjacent pixel transitions:
\begin{equation}
\operatorname{BoundaryLen}(\mathbf{Y})=
\sum_{h=1}^{H}\sum_{w=1}^{W-1}\mathbb{1}[Y(h,w)\neq Y(h,w+1)]
+
\sum_{h=1}^{H-1}\sum_{w=1}^{W}\mathbb{1}[Y(h,w)\neq Y(h+1,w)],
\end{equation}
where $\mathbf{Y}\in\{0,\dots,K-1\}^{H\times W}$ is any aligned segmentation map. The first term counts horizontal label transitions between neighboring azimuthal pixels, and the second term counts vertical label transitions between neighboring depth pixels. Hence, $\operatorname{BoundaryLen}(\mathbf{Y})$ measures the total complexity of the segmentation boundaries in the interval. The topological fragmentation was quantified by the total number of 8-connected components\cite{rosenfeld1966} of each class after discarding components smaller than 10 pixels:
\begin{equation}
\operatorname{TotalComponents}(\mathbf{Y})
=
\sum_{k=0}^{K-1}
\left|
\mathcal{CC}^{(8)}_{\ge 10}\!\left(\mathbb{1}[Y=k]\right)
\right|.
\end{equation}
Here, $\mathbb{1}[Y=k]$ is the binary mask of class $k$, $\mathcal{CC}^{(8)}_{\ge 10}$ denotes the set of 8-connected components with size at least 10 pixels, and $|\cdot|$ denotes the number of such components. Thus, $\operatorname{TotalComponents}(\mathbf{Y})$ summarizes how fragmented the segmentation is across all classes.

Additional selected-case structural diagnostics, including boundary length, connected-component counts, and low-confidence update fractions, were used only for supplementary case-level interpretation. For qualitative comparison, remapped segmentations were also overlaid on the original acoustic image after alignment, and aligned confusion matrices were visualized to assess whether gains were distributed across the full four-class partition rather than concentrated in a single dominant class.

\section*{Acknowledgements}
This work was supported by the Brazilian National Council for Scientific and Technological Development (CNPq) under grant No. 445344/2024-5. The author acknowledges financial support as a Project Coordinator and Researcher Level 1 through the \textit{Talentos Brasil} (Talents Brazil) Fellowship. Furthermore, the author gratefully acknowledges the Group of Artificial Intelligence in Applied Geophysics (GAIA) at the Federal University of Bahia (UFBA) for providing essential computational resources and infrastructure.

\section*{Author contributions statement}
J.L.L.J.S. is the sole author of this work. J.L.L.J.S. conceived the experiments, developed the mathematical formulation and implemented the software framework for the CG-DCA model and the baseline refinement families, conducted all benchmarking and ablation studies, analyzed the results, and wrote and reviewed the manuscript.

\section*{Data availability statement}
The dataset analyzed in this study is the Wellbore Acoustic Image Database (WAID), a public resource provided by Petrobras to promote machine learning applications in the oil and gas industry. The dataset consists of acoustic image logs from several wells in Brazilian pre-salt carbonate reservoirs, alongside associated conventional open-hole logs. The data is publicly and freely available for research purposes through the official Petrobras repository at \url{https://github.com/petrobras/waid}.

\section*{Additional information}
\textbf{Competing interests:} The author declares no competing interests.

\section*{Author note}
This work is currently under peer review. The content of this manuscript may change in a revised or published version.

\section*{Corresponding Author}
Correspondence and requests for materials should be addressed to \texttt{jseluis.silva@gmail.com}.

\bibliography{sample}
\end{document}